%% file: main.tex
\documentclass[10pt,twocolumn,letterpaper]{article}

\usepackage{wacv}
\usepackage{times}
\usepackage{epsfig}
\usepackage{graphicx}
\usepackage{amsmath}
\usepackage{amssymb}
\usepackage{booktabs}
\usepackage{comment}
\usepackage{xcolor}
\usepackage{color}
\usepackage{xspace}
\usepackage{caption}
\usepackage{subcaption}
\usepackage{float}
\usepackage{multirow}
\usepackage[linesnumbered,ruled,vlined]{algorithm2e}
\usepackage{listings}
\usepackage{float}

\def \OURMETHOD {IIVCL\xspace}
\def \BASELINE {\emph{$\rho$-MoCo}\xspace} 

\def\wacvPaperID{} 

\wacvfinalcopy 

\usepackage{hyperref}
\hypersetup{breaklinks=true,colorlinks,bookmarks=false]}

\begin{document}

\title{Nearest-Neighbor Inter-Intra Contrastive Learning from Unlabeled Videos}

\author{David Fan \and
Deyu Yang \and
Xinyu Li \and
Vimal Bhat \and
Rohith MV \and
Amazon Prime Video \\
{\tt\small \{fandavi, deyu, xxnl, vimalb, kurohith\}@amazon.com}
}

\maketitle
\thispagestyle{empty}

\input{main/1_abstract}
\input{main/2_introduction}
\input{main/3_related_work}
\input{main/4_method}
\input{main/5_experiments}
\input{main/6_conclusion}

{\small
\bibliographystyle{ieee_fullname}
\bibliography{main}
}

\newpage

\appendix
\include{supplement}

\end{document}

%% file: main/1_abstract.tex
\begin{abstract}
\noindent Contrastive learning has recently narrowed the gap between self-supervised and supervised methods in image and video domain. State-of-the-art video contrastive learning methods such as CVRL and $\rho$-MoCo spatiotemporally augment two clips from the same video as positives. By only sampling positive clips locally from a single video, these methods neglect other semantically related videos that can also be useful. To address this limitation, we leverage nearest-neighbor videos from the global space as additional positive pairs, thus improving positive key diversity and introducing a more relaxed notion of similarity that extends beyond video and even class boundaries. Our method, Inter-Intra Video Contrastive Learning (IIVCL), improves performance on a range of video tasks.
\end{abstract}

%% file: main/2_introduction.tex
\section{Introduction}
\label{sec:introduction}
The success of supervised deep learning for computer vision can largely be attributed to the availability of large-scale labeled datasets~\cite{zhu2020comprehensive,deng2009imagenet,kay2017kinetics,chen2020oasis,huang2020movienet}, which are difficult and expensive to create. However, progress in compute and model representation capabilities has outpaced progress in dataset creation~\cite{sun2017revisiting,chen2020oasis}. Self-supervised learning is the paradigm of learning from \emph{unlabeled} data to decouple reliance upon large-scale labeled datasets. It has already shown great potential in NLP~\cite{devlin2018bert,brown2020language} for producing representations that generalize well to multiple tasks. 

\begin{figure}[!tb]
	\centering
	\includegraphics[width=0.95\linewidth]{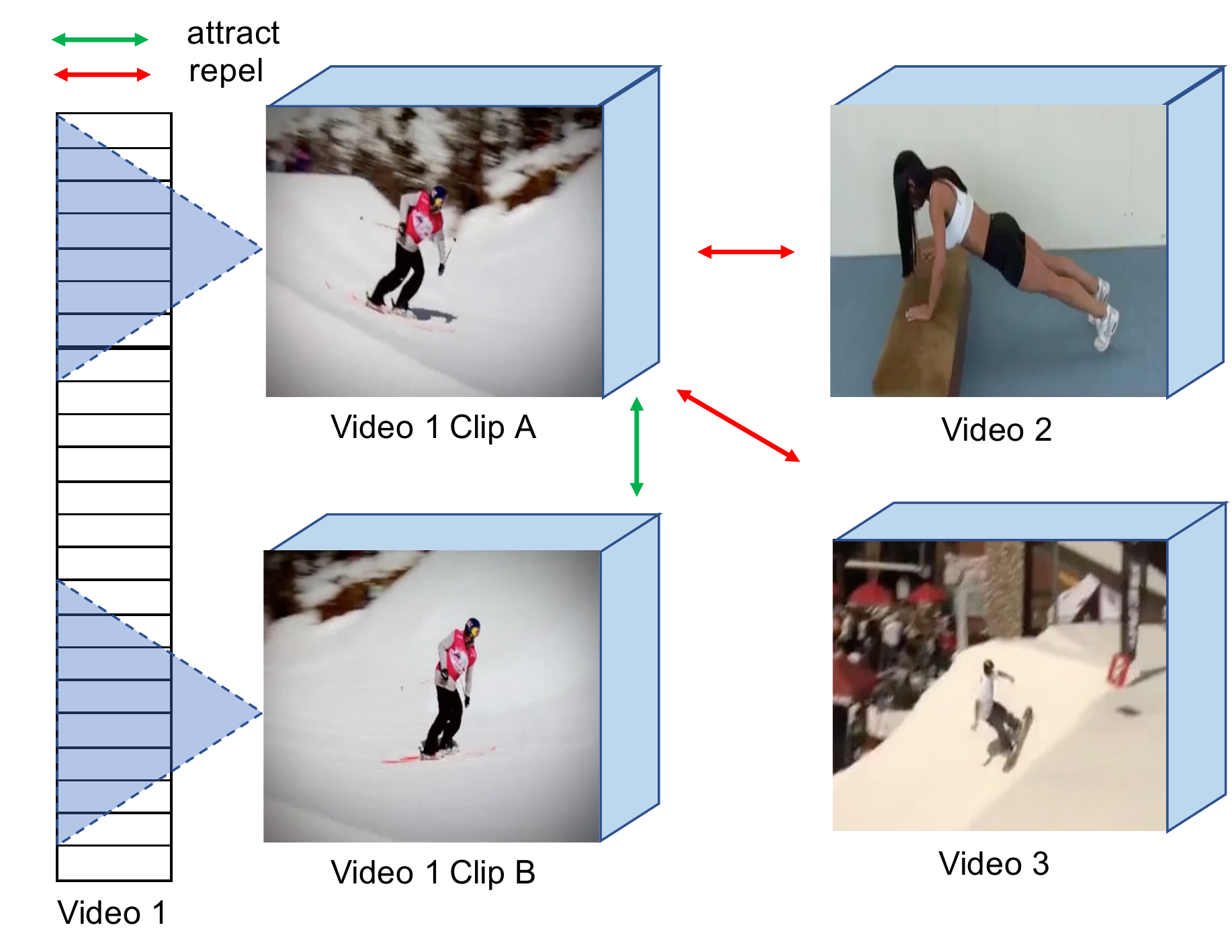}
	\caption{\small{Video contrastive learning methods such as CVRL~\cite{qian2021spatiotemporal} and $\rho$-MoCo~\cite{feichtenhofer2021large} only sample positive pairs within the same video boundary, such as clips A and B. However, this means that different \emph{yet similar} videos such as 3 are \textbf{never} used as positives, even if they are semantically similar to video 1. Our work leverages semantically similar inter-video pairs as positives (e.g. video 3), in addition to intra-video subclips (e.g. 1.A and 1.B) used by previous works. ``Clip'' refers to a subset of video frames.}}
	\label{fig:limitation_example}
\end{figure}
\addtolength{\textfloatsep}{-0.3cm}

Recently, a flavor of self-supervised learning known as contrastive learning / instance discrimination~\cite{oord2018representation} has become dominant in vision through works such as MoCo~\cite{he2020momentum,chen2020improved} and SimCLR~\cite{chen2020simple,chen2020big} which are competitive with supervised learning in image domain. More recently, contrastive learning works for video such as CVRL~\cite{qian2021spatiotemporal} and $\rho$-MoCo~\cite{feichtenhofer2021large} are competitive with supervised learning. These works learn a representation from unlabeled data by pulling positive pairs closer and pushing negative samples apart in the embedding space. In the case of video, these positive pairs are generated through random augmentations of sub-clips from the \emph{same} video~\cite{qian2021spatiotemporal,feichtenhofer2021large}, while clips from other similar videos are \emph{never} used as positives.

By only considering clips that belong to the same video to be positive, works such as CVRL~\cite{qian2021spatiotemporal} and $\rho$-MoCo~\cite{feichtenhofer2021large} neglect other semantically related videos that may also be useful and relevant as positives for contrastive learning. Fig.~\ref{fig:limitation_example} depicts such a scenario. Clips A and B from the same skiing video are the positive pair while video 2 (push-up) and video 3 (snowboarding) are negatives. Clips A and B are certainly more similar to each other than to the push-up and snowboarding videos, but the snowboarding video is far more similar to the positive anchor when compared to an indoor activity such as doing push-ups. Yet the relative similarity between skiing - snowboarding will never be exploited by methods such as CVRL~\cite{qian2021spatiotemporal} and $\rho$-MoCo~\cite{feichtenhofer2021large}.

\begin{figure*}[!t]
	\centering

	\begin{subfigure}[t]{0.32\textwidth}
		\centering
		\includegraphics[width=0.9\linewidth]{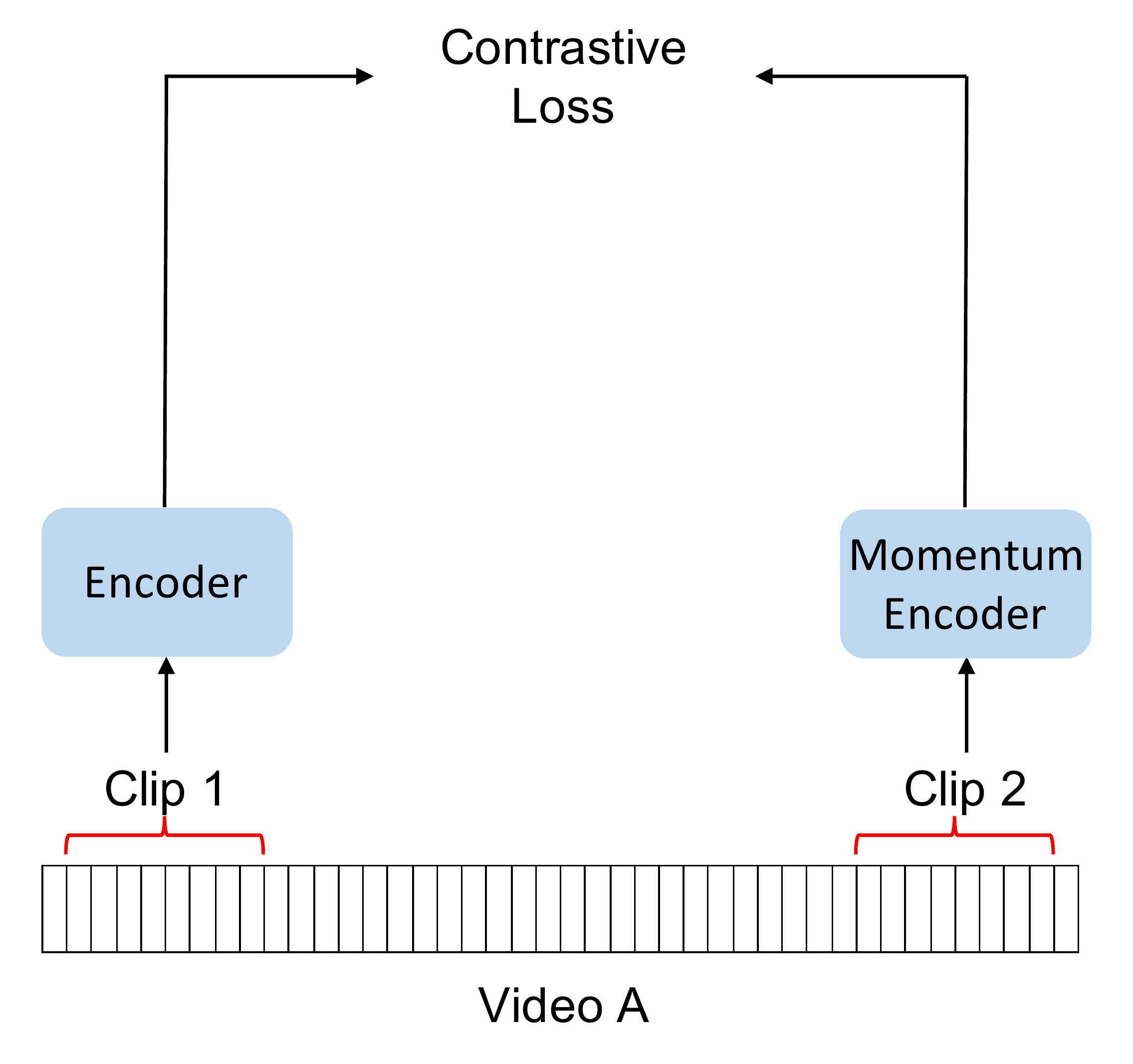}
		\caption{\small{Intra-video pair task.~\cite{qian2021spatiotemporal,feichtenhofer2021large}}}
		\label{fig:algorithm_overview:spatiotemporal}
	\end{subfigure}
	~
	\begin{subfigure}[t]{0.32\textwidth}
		\centering
		\includegraphics[width=0.9\linewidth]{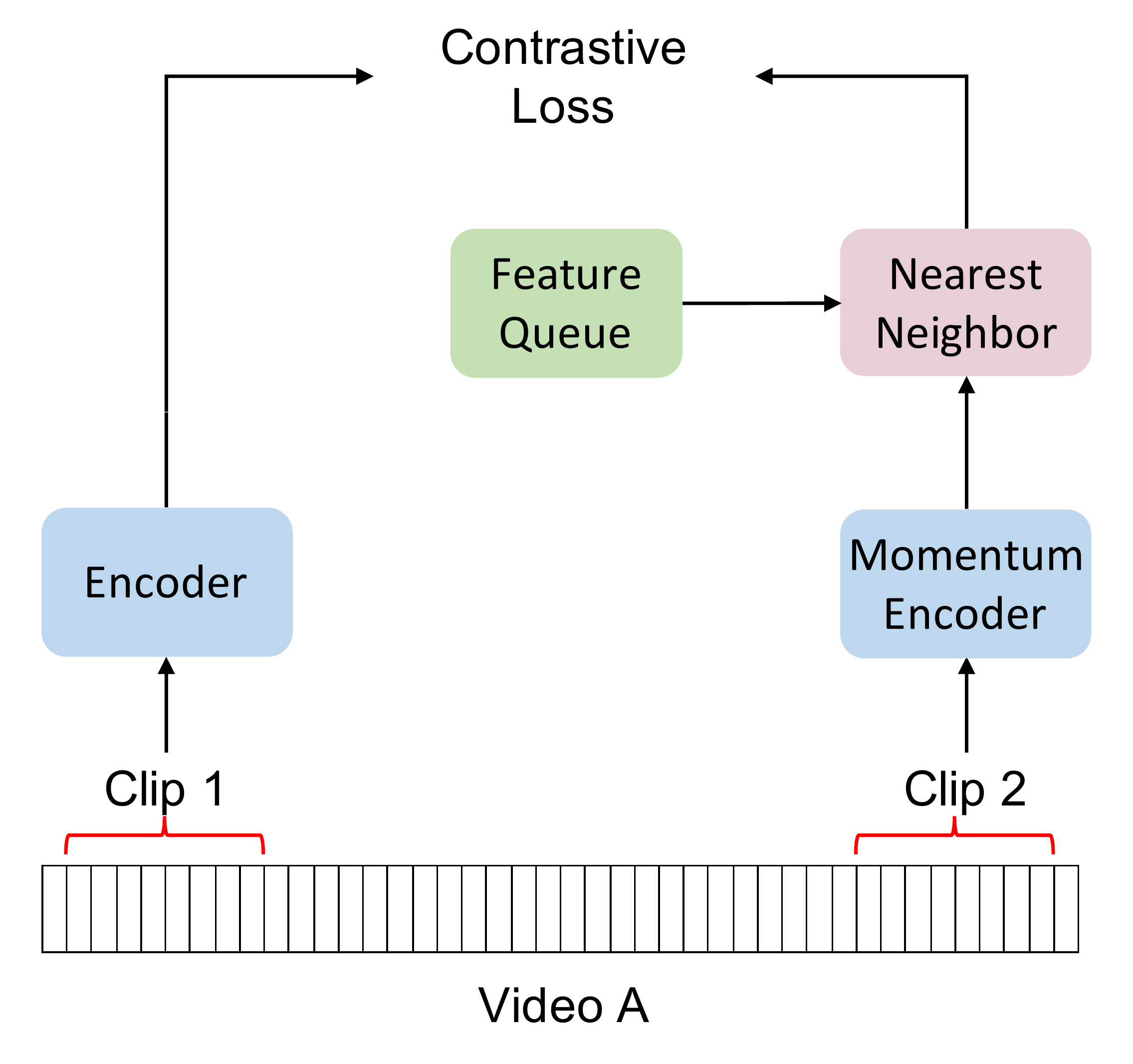}
		\caption{\small{Inter-video nearest-neighbor task.}}
		\label{fig:algorithm_overview:nearest_neighbor}
	\end{subfigure}
	~
	\begin{subfigure}[t]{0.32\textwidth}
		\centering
		\includegraphics[width=0.9\linewidth]{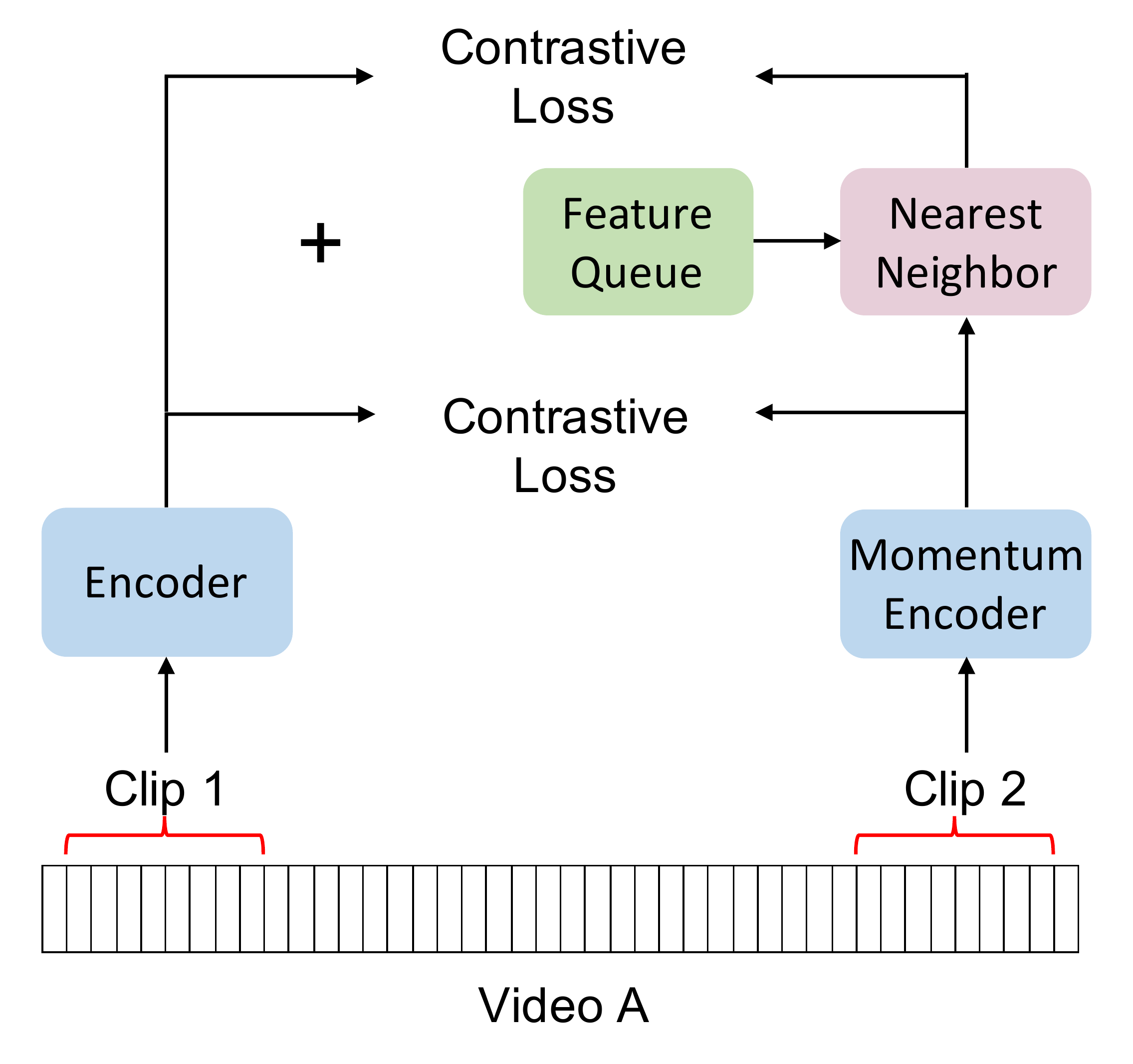}
		\caption{\small{Inter-Intra video contrastive learning.}}
		\label{fig:algorithm_overview:combined}
	\end{subfigure}

	\caption{\small{\OURMETHOD{} Overview. Fig.~\ref{fig:algorithm_overview:spatiotemporal} shows intra-video positive sampling as used by current state-of-the-art~\cite{qian2021spatiotemporal,feichtenhofer2021large}. Fig.~\ref{fig:algorithm_overview:nearest_neighbor} shows our proposal to leverage nearest-neighboring samples from an evolving queue as positives for contrastive learning. Fig.~\ref{fig:algorithm_overview:combined} is our \OURMETHOD{} which combines a) and b) to learn similarity both within the same video and between different videos. NN sampling is simple, light-weight, and directly plugs into contrastive frameworks. Note that there is no temporal ordering between clips 1 and 2 as they are randomly sampled.}}
	\label{fig:algorithm_overview}
\end{figure*}

This raises the question of what constitutes a desirable video representation; by focusing too much on local intra-video semantics, we may miss the larger picture and hierarchy of visual concepts. This might lead to overfitting to tasks that are similar to the pretraining dataset and thus hurt generalization. On the other hand, if we focus too much on global inter-video semantics, we may lose sight of granular details that are also important for video understanding.

To balance the two, we propose learning notions of similarity both within the same video and between different videos, by leveraging inter-video nearest-neighbor (NNs) from the global space in addition to existing intra-video clips as positive pairs for contrastive learning. For any pair of intra-video clips, our method ``Inter-Intra Video Contrastive Learning'' (\OURMETHOD{}) defines a second positive key as the most similar video found from a dynamically evolving queue of randomly sampled videos in the learned representation space, as shown in Fig.~\ref{fig:algorithm_overview}.

\begin{figure*}[!tb]
	\centering

	\begin{subfigure}[t]{0.48\textwidth}
		\centering
		\includegraphics[width=0.95\linewidth]{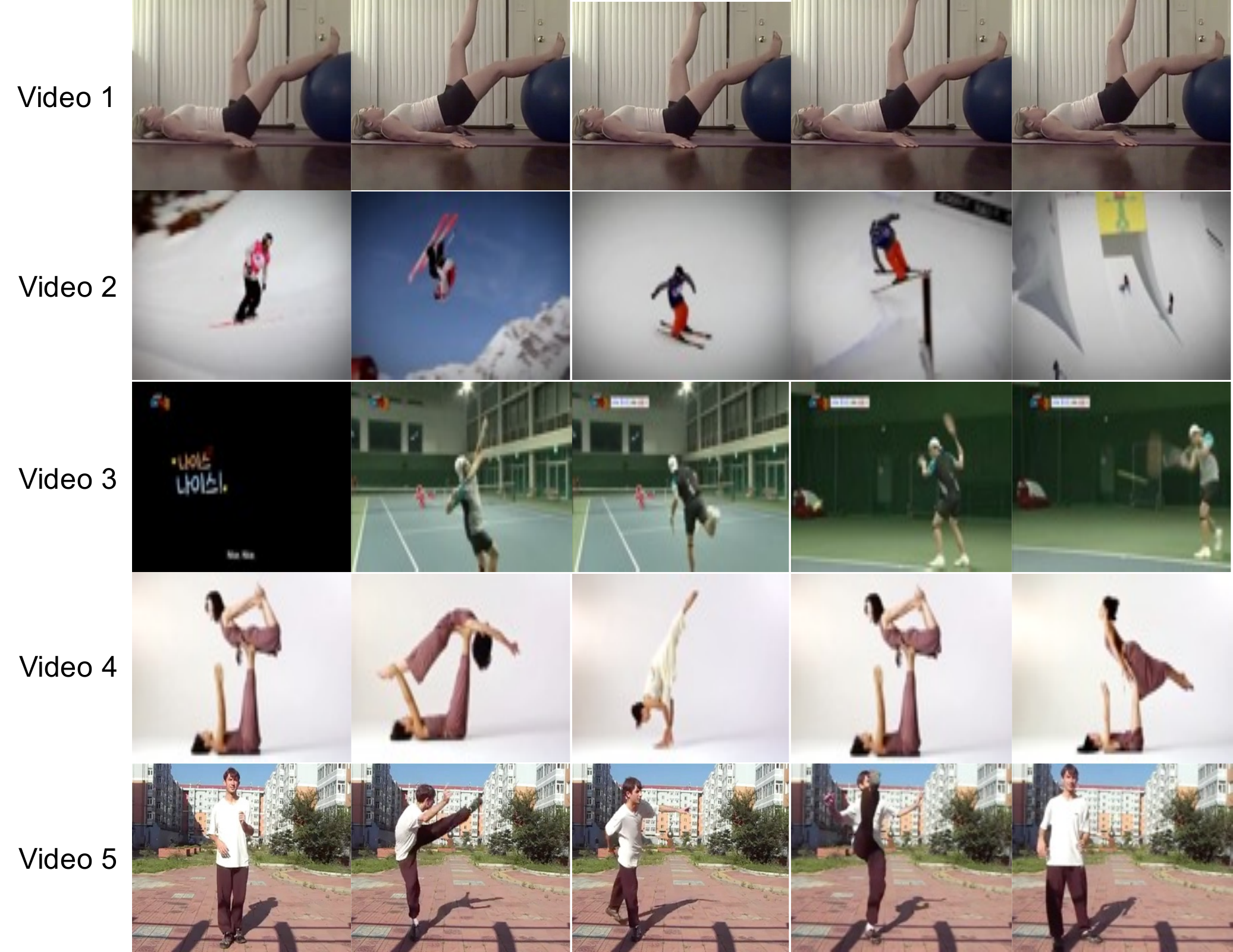}
		\caption{\footnotesize{Intra-video sampling results in low diversity of positive keys because subclips belong to the same short video. Each row shows a different video.}}
		\label{fig:nearest_neighbor_evolution:intravideo}
	\end{subfigure}
	~
	\begin{subfigure}[t]{0.48\textwidth}
		\centering
		\includegraphics[width=0.95\linewidth]{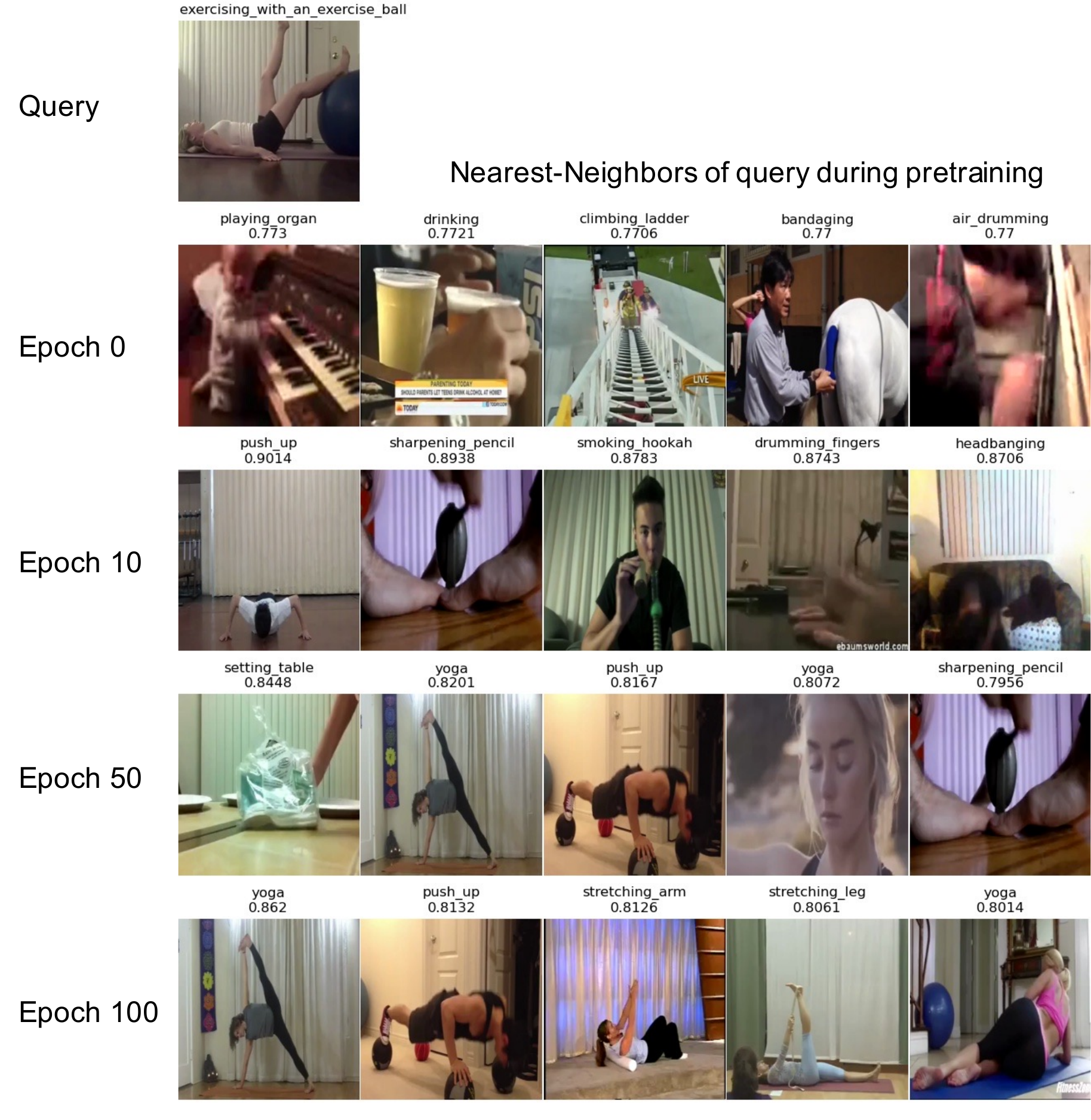}
		\caption{\footnotesize{For a query image (first row), the evolution of the top-5 nearest-neighbors during SSL pretraining starting from random initialization.}}
		\label{fig:nearest_neighbor_evolution:exercise_ball}
	\end{subfigure}

	\caption{\small{Intra-video vs. inter-video nearest-neighbor positives. In a), positive keys are always restricted to a single video boundary, which limits diversity. In b), positive keys are sampled globally using similarity in the learned feature space, which improves during pretraining. NNs do not necessarily belong to the same semantic class as the query. This global notion of similarity improves generalization.}}
	\label{fig:nearest_neighbor_evolution}
\end{figure*}

The benefit of adding globally sampled NNs as positives is two-fold. First, NNs present additional diversity in viewpoint and scene orientation, which cannot be expressed through sampling subclips from the \emph{same} video, as shown by comparing Fig.~\ref{fig:nearest_neighbor_evolution} a) and b). This is true whether the subclips are augmented, shuffled or otherwise modified per the pretext task, so long as they belong to the same video. Image works such as NNCLR~\cite{Dwibedi_2021_ICCV} show that improving positive key diversity leads to better generalization on image tasks, and we explore whether the same is true for video.

Second, because the set of positive keys is expanded beyond the boundaries of a single video, it becomes possible to also learn from weaker yet relevant and useful positive pairs, such as the snowboarding-skiing example from Fig.~\ref{fig:limitation_example}. By combining these nearest-neighbor pairs with intra-video pairs as positives for contrastive learning, the model can learn notions of similarity both within the same video and between different videos, thus striking a balance between learning granular details and learning higher-level visual concepts. We then evaluate whether this leads to improved representations on downstream video tasks.

In summary, our contributions are the following:

(i) Unlike other video-based works, we go beyond single-video positives using only RGB by leveraging globally sampled nearest-neighboring videos as a simple and effective way to increase the semantic diversity of positive keys and introduce higher-level notions of similarity.

(ii) We introduce \OURMETHOD{} as a novel yet \emph{simple} self-supervised video representation learning method that learns a joint representation of both intra-video pairs and nearest-neighbors --- without clustering nor multiple modalities.

(iii) We demonstrate that striking a balance between local and global similarity leads to improved performance over existing intra-video contrastive learning methods~\cite{qian2021spatiotemporal,feichtenhofer2021large} on video action recognition, action detection, and video retrieval --- even in a few-shot learning setting.

%% file: main/3_related_work.tex
\section{Related Work}
\label{sec:relatedwork}

\noindent \textbf{Self-supervised image representation learning.} \quad Earlier works focused on designing pretext tasks whose solution would yield a useful representation for downstream classification. Some pretext tasks include predicting the relative positioning of patches~\cite{doersch2015unsupervised}, image rotation~\cite{gidaris2018unsupervised}, image colorization~\cite{zhang2016colorful}, solving jigsaw puzzles~\cite{noroozi2016unsupervised}, and counting visual primitives~\cite{noroozi2017representation}. While promising, these approaches were not competitive with fully-supervised representations~\cite{kolesnikov2019revisiting,newell2020useful}, partly because it is easy to learn shortcuts that trivially solve the pretext task. 

The re-emergence of contrastive learning elevated self-supervised image representation learning as a viable alternative paradigm to fully-supervised learning~\cite{he2020momentum,chen2020improved,chen2020simple,chen2020big,grill2020bootstrap,chen2021exploring}. These methods encourage models to be invariant to multiple augmented views of the same image.~\cite{tian2020contrastive} generates multiple views from the same image by splitting images into luminance and chrominance space.

Some recent works such as~\cite{zhuang2019local,caron2020unsupervised,li2020prototypical} go beyond single-instance contrastive learning by using clusters to find semantically relevant positive pairs. Our work is similar to NNCLR~\cite{Dwibedi_2021_ICCV} which uses nearest-neighbors for image contrastive learning, but differs in that we combine inter-video nearest-neighbors with intra-video positives for \textbf{video} contrastive learning, which brings unique challenges.

\noindent \textbf{Self-supervised video representation learning.} \quad Temporal information in videos enables the design of interesting pretext tasks that are not possible with images. Video in general is more challenging due to its temporal dimension. Pretext tasks for video self-supervised learning include predicting future frames~\cite{diba2019dynamonet,luo2017unsupervised,vondrick2016anticipating,srivastava2015unsupervised}, the correct order of shuffled frames~\cite{misra2016shuffle,fernando2017self} or shuffled clips~\cite{lee2017unsupervised}, video rotation~\cite{jing2018self}, playback speed~\cite{benaim2020speednet}, directionality of video~\cite{wei2018learning}, motion and appearance statistics~\cite{wang2019self}, and solving space time cubic puzzles~\cite{kim2019self}. Other works have leveraged video correspondence by tracking patches~\cite{wang2015unsupervised}, pixels~\cite{wang2019learning}, and objects~\cite{jabri2020space} across adjacent frames. 

Several recent works have considered contrastive learning for video. These methods differ in their definition of positive and negative samples. Some works use different subclips of equal-length from the same video as positive samples~\cite{qian2021spatiotemporal,feichtenhofer2021large,dave2021tclr}, or different frames from the same video~\cite{tian2020contrastive}.~\cite{recasens2021broaden} samples two subclips of different length from the same video to encourage generalization to broader context. Other works utilize optical flow in a cross-modal context;~\cite{han2020self} uses optical flow to mine RGB images with similar motion cues as positives, while other works do not mine positives but instead learn from the natural correspondence between optical flow and RGB within the same video~\cite{xiao2021modist}, or within the same frame~\cite{tian2020contrastive}. Another set of works utilize pretext tasks such as pace prediction~\cite{yang2020video,wang2020self}, clip shuffling~\cite{xu2019self}, or a combination of both~\cite{kuang2021video}.

In contrast, our work goes beyond local definitions of positives from a single-video and expands to globally sampled nearest-neighbor videos, but without using optical flow nor separate training phases like~\cite{han2020self}. Unlike works such as~\cite{chen2021multimodal}, our work uses the online representation space to pick NNs on the fly instead of pre-computing video clusters. Unlike methods such as~\cite{chen2021multimodal,morgado2021audio} that go beyond single-video positives, our work only uses RGB frames.

%% file: main/4_method.tex
\section{Inter-Intra Video Contrastive Learning}
\label{sec:method}

\subsection{Contrastive loss}
Contrastive learning maximizes the similarity of a given embedded sample $q$ with its embedded positive key $k^+$, while minimizing similarity to negative embeddings $n_i$. In the rest of this work, we refer to ($q$, $k^+$) as ``positive pairs''. We utilize the InfoNCE loss~\cite{oord2018representation} for self-supervised video representation learning, which is given below:
\begin{equation}
\label{infonce_loss}
\mathcal{L}^{\textrm{NCE}}(q, k^+,\ \mathcal{N}^-) = -\textrm{log} \frac{\textrm{exp}(sim(q, k^+) / \tau)}{\sum\limits_{k \in \{k^+\} \cup \mathcal{N}^-}\textrm{exp}(sim(q, k) / \tau)}
\end{equation}
\noindent where $\tau > 0$ is a temperature hyper-parameter and $sim(\cdot)$ denotes the similarity function --- which in this work is the dot product (cosine) similarity between two $\ell_2$ normalized vectors: $sim(q, k) = q \cdot k = q^{T}k / (||q|| \ ||k||)$. 

\subsection{Multi-Task Objective}
The manner in which positive pairs are sampled during contrastive learning introduces an inductive bias that has a significant impact on downstream task performance~\cite{xiao2020should}. For example, works such as MoCo~\cite{he2020momentum,chen2020improved} and SimCLR~\cite{chen2020simple,chen2020big} that use image augmentation encourage model invariance to different colors, crops, and orientations, potentially at the expense of other inductive bias. \newline

\noindent \textbf{Intra-Video Contrastive Learning.} Contrastive learning methods for video such as CVRL~\cite{qian2021spatiotemporal} and $\rho$-MoCo~\cite{feichtenhofer2021large} use the embeddings of two subclips $z_1$ and $z_2$ from the same video as positives. Intra-video positive sampling teaches the model to identify whether two different clips correspond to the same video. Given a queue $Q$ of randomly sampled embeddings, the intra-video contrastive loss is then:
\begin{equation}
\label{intravideo_loss}
\mathcal{L}_{\textrm{\emph{Intra}}}(z_1, z_2,\ Q) = \mathcal{L}^{\textrm{NCE}}(z_1, z_2,\ Q)
\end{equation}

\noindent \textbf{Nearest-Neighbor Contrastive Learning.}
Motivated by the observation in Fig.~\ref{fig:limitation_example} that intra-video sampling excludes other semantically similar videos from ever being used as positives; we seek to expand the positive keyset beyond individual video boundaries to improve diversity. Offline positive sets are not suitable because they cannot be efficiently updated at every iteration or even at every epoch. Clustering to find similar videos requires additional hyperparameters. Thus, we maintain a queue $Q$ that is updated with embeddings from each forward pass (design details in Sec.~\ref{sec:method:pretraining_details}), which allows us to directly compute cosine similarities between the input video and queue. Given an embedded input video $x$ and queue $Q$ of randomly sampled embeddings across the dataset, we use the nearest-neighbor of $x$ as its positive key:
\begin{equation}
\label{nearest_neighbor_eq}
\textrm{NN}(x, Q) = \mathop{\mathrm{argmax}}_{z \in Q} (x \cdot z)
\end{equation}

\noindent Let $z_1$ and $z_2$ be the embeddings of two subclips from the same video. We use $Q$ for both selecting the NN as a positive key and providing negatives (excluding the NN). Using the nearest-neighbor operation in Eq.~\ref{nearest_neighbor_eq} to select the positive key for $z_1$ as $\textrm{NN}(z_2, Q)$, and removing it from $Q$ to form $Q^- = Q \setminus \textrm{NN}(z_2, Q)$, we have the NN contrastive loss:
\begin{equation}
    \label{intervideo_loss}
    \mathcal{L}_{\textrm{\emph{NN}}}(z_1, z_2,\ Q) = \mathcal{L}^{\textrm{NCE}}(z_1, \textrm{NN}(z_2, Q),\ Q^{-})
\end{equation}

\noindent \textbf{Combined Intra and Inter Training Objective} \newline
We use the same backbone but separate MLP projection heads to process the intra-video and NN positive pairs. As each of these pretext tasks learns a different notion of similarity, we combine them via a multi-task loss. We also maintain two separate queues of embeddings: $Q_\textrm{Intra}$ and $Q_\textrm{NN}$. $Q_\textrm{NN}$ is used both to find the NN and provide negative keys (excluding the NN), while $Q_\textrm{Intra}$ only provides negative keys. We expand on these details in section~\ref{sec:method:pretraining_details}.

Specifically, let $f^q(\cdot)$ and $f^k(\cdot)$ be the encoder and its offline momentum-updated version, $g^{\textrm{Intra}}(\cdot)$ and $g^{\textrm{NN}}(\cdot)$ be two separate MLP heads, and $x_1$ and $x_2$ be two subclips sampled from the same video. We first obtain the embeddings ($z_1^{\textrm{Intra}}$, $z_2^{\textrm{Intra}}$) and ($z_1^{\textrm{NN}}$, $z_2^{\textrm{NN}}$).
\begin{align*}
    & z_1^{\textrm{Intra}} = g^{\textrm{Intra}}(f^{q}(x_1)); \quad z_2^{\textrm{Intra}} = g^{\textrm{Intra}}(f^{k}(x_2)) \\
    & z_1^{\textrm{NN}} = g^{\textrm{NN}}(f^{q}(x_1)); \quad z_2^{\textrm{NN}} = g^{\textrm{NN}}(f^{k}(x_2))
\end{align*}
After obtaining the embeddings, we combine Eqs.~\ref{intravideo_loss} and~\ref{intervideo_loss} to get the final training objective. Note that we use a symmetric loss but show only one side for simplicity. $\lambda_{\textrm{\emph{Intra}}}$ and $\lambda_{\textrm{\emph{NN}}}$ are tunable parameters that control the contribution of each loss, which in our work is 1.0 for both. An ablation for this is in Tab. \ref{table:nn_weight}.
\begin{equation}
\begin{aligned}
\label{combined_loss}
\mathcal{L}(z_1^{\textrm{Intra}}, z_2^{\textrm{Intra}}, z_1^{\textrm{NN}}, z_2^\textrm{NN}) &= \lambda_{\textrm{\emph{Intra}}} \cdot  \mathcal{L}_{\textrm{\emph{Intra}}}(z_1^{\textrm{Intra}}, z_2^{\textrm{Intra}}, Q_\textrm{Intra}) \\
&+ \lambda_{\textrm{\emph{NN}}} \cdot \mathcal{L}_{\textrm{\emph{NN}}}(z_1^{\textrm{NN}}, z_2^{\textrm{NN}}, Q_\textrm{NN})
\end{aligned}
\end{equation}
Note that class labels are not used and the model is free to learn its own notion of similarity. Over the course of pretraining, the model becomes better at picking semantically similar nearest-neighbor video clips while introducing additional diversity of positive samples that is not possible through sampling intra-video clips, as shown in Fig.~\ref{fig:nearest_neighbor_evolution}.

\subsection{Pretraining Methodology}
\label{sec:method:pretraining_details}
\subsubsection{Momentum Encoder and Queue}
We make several design choices that enable end-to-end learning from unlabeled videos using our method. As contrastive learning requires large batch sizes~\cite{chen2020simple} and computing video embeddings is expensive, we use a FIFO queue that is updated with embeddings from a momentum encoder, similar to~\cite{he2020momentum,chen2020improved}. The momentum encoder's weights $\theta_k$ are updated as a moving average of the encoder's weights $\theta_q$, with momentum coeff. $m \in [0, 1)$, as given by Eq.~\ref{momentum_update_eq}. Thus the momentum encoder receives no gradients.
\begin{equation}
    \label{momentum_update_eq}
    \theta_k \leftarrow m \theta_k + (1 - m) \theta_q
\end{equation}

We share the same encoder but utilize separate MLP heads when processing intra-video and nearest-neighbor positives. We also maintain two separate queues for each task. Note that the queue contains approximate representations of a large subset of pretraining data in memory and is dynamically updated ``for free'' since we only use embeddings that are already computed during the forward pass. Our method scales to large data and is more efficient than methods that utilize clustering such as~\cite{caron2020unsupervised,li2020prototypical,chen2021multimodal,patrick2020multi}. Our method also allows for more up-to-date representations than methods that use an offline positive set~\cite{morgado2021audio,chen2021shot}.

\vspace{-0.2cm} \subsubsection{Implementation Details}
\label{sec:method:other_details}
\noindent \textbf{Loss.} \quad
The temperature $\tau$ = 0.1. $\lambda_{\textrm{\emph{Intra}}}$ = 1.0, $\lambda_{\textrm{\emph{NN}}}$ = 1.0.

\noindent \textbf{Encoder.} \quad
For all experiments, we use a ResNet3D-50 8x8 slow pathway from~\cite{feichtenhofer2019slowfast} with He initialization~\cite{he2015delving}, unless otherwise indicated. Outputs are taken after the global average pooling layer to form a 2048-d embedding. Following~\cite{chen2020improved,feichtenhofer2021large}, we use a 3-layer projection MLP during pretraining only. The MLP has hidden dimension 2048 and final embedding dimension of 128 with no batch norm (BN). The MLP is then removed for downstream experiments. As mentioned above, we use two separate MLP heads for producing intra-video and NN embeddings.

\noindent \textbf{Pretraining Hyperparameters.} \quad
We train for 200 epochs using SGD optimizer (momentum 0.9, weight decay $10^{-4}$) with a total batch size of 512. BN statistics are computed per GPU. We linearly warmup the learning rate to $0.4$ over the first 35 epochs, then use half-period cosine decay~\cite{loshchilov2016sgdr}.

We use a queue storing 65536 negatives and shuffling-BN to avoid information leakage and overfitting~\cite{he2020momentum}. The momentum encoder weights are updated per Eq.~\ref{momentum_update_eq} with an annealed momentum coeff. as in~\cite{feichtenhofer2021large}, initialized to 0.994.

\noindent \textbf{Data and Augmentations}
We sample two 8-frame clips with a temporal stride of 8 from each video for self-supervised pretraining. We apply random shortest-side resizing to [256, 320] pixels, color jittering (ratio 0.4, p=0.8), grayscale conversion (p=0.2), Gaussian blur (p=0.5), horizontal flip (p=0.2), and random cropping to 224 $\times$ 224.

%% file: main/5_experiments.tex
\section{Experiments}
\label{sec:experiments}
\subsection{Baselines}
\label{sec:experiments:baseline}
CVRL \cite{qian2021spatiotemporal} and $\rho$-MoCo \cite{feichtenhofer2021large} are two leading contrastive learning works that sample intra-video clips. We primarily compare against $\rho$-MoCo for $\rho$=2 (two clips per video), pretrained for 200 epochs on unlabeled K400, and call this baseline  \BASELINE{}. The original paper does not test \BASELINE{} on all downstream datasets, so we rerun all downstream experiments (even those reported in the paper) for fair comparison. We compare our \OURMETHOD{} against \BASELINE{} to distill the effect of improved positive key diversity and balanced global-local context on downstream task performance.

\begin{table*}[!t]
  \footnotesize
\begin{center}
  \begin{tabular}{lc|clcc|c|c|c}
           &  &  & Pretrain & Pretrain & Pretrain & & \\
          Method & Date & Backbone & Data (duration) & Epochs & Input Size & UCF & HMDB & K400 \\
          \hline
          \textcolor{gray}{Supervised} & & \textcolor{gray}{R3D-50} & \textcolor{gray}{scratch} & & \textcolor{gray}{$8 \times 224^2$} & \textcolor{gray}{68.8} & \textcolor{gray}{22.7} & \textcolor{gray}{74.7} \\
          \hline
          DPC~\cite{han2019video} & 2019 & R2D-3D34 & K400 (28d) & 110 & $40 \times 224^2$ & 75.7 & 35.7 & - \\
          CBT~\cite{sun1906contrastive} & 2019 & S3D & K600+(273d) & 130 & $112^2$ & 79.5 & 44.6 & - \\
          DynamoNet~\cite{diba2019dynamonet} & 2019 & STCNet & YT8M-1 (58d) & - & $32 \times 112^2$ & 88.1 & 59.5 & - \\
          SpeedNet~\cite{benaim2020speednet} & 2020 & S3D-G & K400 (28d) & - & $16 \times 224^2$ & 81.1 & 48.8 & - \\
          MemDPC~\cite{han2020memory} & 2020 & R2D-3D34 & K400 (28d) & - & $40 \times 224^2$ & 86.1 & 54.5 & - \\
          VideoMoCo~\cite{pan2021videomoco} & 2021 & R(2+1)D18 & K400 (28d) & 200 & $32 \times 224^2$ & 78.7 & 49.2 & - \\
          TCLR~\cite{dave2021tclr} & 2021 & R(2+1)D18 & K400 (28d) & 400 & $16 \times 112^2$ & 84.1 & 53.6 & - \\
          VCLR~\cite{kuang2021video} & 2021 & R2D-50 & K400 (28d) & 400 & $32 \times 224^2$ & 85.6 & 54.1 & 64.1 \\
          LSFD~\cite{behrmann2021long} & 2021 & R3D-18 & K400 (28d) & 500 & $16 \times 224^2$ & 77.2 & 53.7 & - \\
          TECVRL~\cite{jenni2021time} & 2021 & R3D-18 & K400 (28d) & 200 & $16 \times 128^2$ & 87.1 & 63.6 & - \\
          \textbf{\OURMETHOD{}} & & R3D-18 & K400 (28d) & 200 & $8 \times 128^2$ & 89.4 & 60.2 & 59.2 \\
          \hline
          VTHCL~\cite{yang2020video} & 2020 & R3D-50 & K400 (28d) & 200 & $8 \times 224^2$ & 82.1 & 49.2 & 37.8 \\
          CVRL~\cite{qian2021spatiotemporal} & 2020 & R3D-50 & K400 (28d) & 1000 & $16 \times 224^2$ & 92.2 & 66.7 & 66.1 \\
        
          $\rho$-MoCo$^{\dagger}$~\cite{feichtenhofer2021large} & 2021 & R3D-50 & K400 (28d) & 200 & $8 \times 224^2$ & 91.1 & 65.3 & 65.4 \\
          $\rho$-MoCo$^{\dagger}$~\cite{feichtenhofer2021large} & 2021 & R3D-50 & K400 (28d) & 400 & $8 \times 224^2$ & 92.5 & - & 67.4 \\
          \textbf{\OURMETHOD{}} & & R3D-50 & K400 (28d) & 200 & $8 \times 224^2$ & 92.6 & 65.8 & 65.7 \\
          \textbf{\OURMETHOD{}} & & R3D-50 & K400 (28d) & 400 & $8 \times 224^2$ & 93.3 & 68.1 & 67.1 \\
          \hline
        \end{tabular}
\end{center}
\caption{\small{\textbf{Comparison with state-of-the-art self-supervised approaches}. Reported results are top-1 accuracy under finetune protocol (UCF, HMDB) and linear protocol (K400). We do not compare against two-stream methods. \\
$^{\dagger}$ refers to our reimplementation (see Sec. \ref{sec:experiments:baseline}).}}
\label{table:sota_comparison}
\end{table*}

\subsection{Action Recognition on UCF101, HMDB51, and Kinetics-400}
\label{sec:action_recognition_ucf_hmdb_k400}
Unless otherwise noted, we train \OURMETHOD{} on unlabeled K400~\cite{kay2017kinetics} (240K videos) for 200 epochs. After self-supervised pretraining, we transfer the learned weights to target datasets for downstream evaluation. We use two evaluation protocols that are popular in the literature for evaluating the quality of self-supervised representations: \textbf{(i) Linear evaluation} freezes the backbone and trains a linear classifier on the target dataset, and \textbf{(ii) Finetuning} trains the entire network end-to-end on the target dataset.

We first report top-1 accuracy on three of the most popular action datasets. For UCF101~\cite{soomro2012ucf101} and HMDB51~\cite{kuehne2011hmdb}, we report finetuning top-1 accuracy on split 1. UCF101 contains 9.5K/3.7K train/test videos with 101 action classes, and HMDB51 contains 3.5K/1.5K videos (mostly from movies) with 51 action classes. For K400~\cite{kay2017kinetics}, we report linear evaluation top-1 accuracy. Kinetics contains 240K/19K train/test videos with 400 action classes. We sample $8 \times 8$ clips for all datasets. At test-time, we use standard 10 (temporal) $\times$ 3 (spatial) crop evaluation~\cite{feichtenhofer2019slowfast}. We report the avg. of three runs.

\vspace{-0.2cm}\subsubsection{Comparison to state-of-the-art.}
In Table \ref{table:sota_comparison}, we compare \OURMETHOD{} against state-of-the-art self-supervised methods in the literature that use only RGB frames (no optical flow nor audio), as well as our intra-video baseline \BASELINE{} using the protocols introduced at the beginning of Sec. \ref{sec:action_recognition_ucf_hmdb_k400}.

First, we compare \OURMETHOD{} against the \BASELINE{} baseline to distill the effect of improved positive key diversity and added global context on downstream performance. \OURMETHOD{} outperforms \BASELINE{} by 1.5\% on UCF, 0.5\% on HMDB, and by 0.3\% on K400. The smaller performance delta on K400 may be due to the fact that linear evaluation is not as reflective of real performance as finetuning the entire network~\cite{newell2020useful}. Consistent improvements over \BASELINE{} highlights the effectiveness of combining intra-video and nearest-neighbor sampling, compared to sampling only intra-video positives.

Next, we compare \OURMETHOD{} against other methods. \OURMETHOD{} also outperforms works that trained for more epochs (e.g. VTHCL \cite{yang2020video}) or used more frames during pretraining (e.g. SpeedNet \cite{benaim2020speednet}, VideoMoCo \cite{pan2021videomoco}). \OURMETHOD{} pretrained for 400 epochs surpasses CVRL~\cite{qian2021spatiotemporal} which pretrained for 2.5x more epochs using 2x more frames.

To fairly compare against other video contrastive learning works, we also present results for a weaker version of our model trained for 200 epochs with a smaller backbone (R18) and smaller input resolution (128x128). Under this setting, our model still convincingly outperforms methods that use larger backbones such as VTHCL~\cite{yang2020video} and MemDPC~\cite{han2020memory}, and larger input resolution such as VideoMoco~\cite{pan2021videomoco} and LSFD~\cite{behrmann2021long}.

\subsection{Action Recognition on Something-Something v2}
To further demonstrate the effectiveness of nearest-neighbor sampling for video contrastive learning, we evaluate \OURMETHOD{} on Something-Something v2~\cite{goyal2017something} (SSv2), which is a challenging benchmark focused on understanding granular motions. Unlike UCF101 and HMDB51 which are very similar to K400, SSv2 distinguishes between directionality for the same higher-level action. For example "putting something \textbf{into} something" vs. "putting something \textbf{next to} something" are different action classes, as well as "moving something up" vs. "pushing something from right to left". We finetune on SSv2 using the same settings as~\cite{feichtenhofer2021large}.

\OURMETHOD{} marginally outperforms the \BASELINE{} baseline. As SSv2 is very different in nature from the K400 dataset which we pretrained on, these results further demonstrate that improving positive key diversity and introducing global similarity leads to more generalized video representations that outperform pure intra-video sampling-based methods.

\begin{table}[!t]
  \footnotesize
  \renewcommand{\arraystretch}{1.1}
\begin{center}
  \begin{tabular}{c|c|c|c}
          Method  & Backbone & Pretrain Data & Top-1 Acc \\
          \hline
          \textcolor{gray}{Supervised} \cite{feichtenhofer2019slowfast} & \textcolor{gray}{R3D-50} & \textcolor{gray}{K400} & \textcolor{gray}{52.8} \\
          \BASELINE{}~\cite{feichtenhofer2021large} & R3D-50 & K400 & 53.6 \\
          \hline
          \textbf{\OURMETHOD{}} & R3D-50 & K400 & 53.8 \\
          \hline
        \end{tabular}
\end{center}
\caption{\small{\textbf{Action recognition on Something-Something.} We finetune on SSv2 using a clip size of $8 \times 8$ and report top-1 accuracy.}}
\label{table:ssv2_eval}
\end{table}

\subsection{Action Detection on AVA}
Thus far, we have demonstrated that \OURMETHOD{} can generalize to new domains within the same task of action recognition. To test whether our method can also generalize to novel downstream tasks, we evaluate \OURMETHOD{} on the new task of action detection, which not only requires accurately classifying the action but also localizing the bounding box of the person performing the action.

We evaluate on the AVA dataset~\cite{gu2018ava} which contains 221K/57K training and validation videos, and report mean Average Precision (mAP) at IOU threshold 0.5. We follow~\cite{feichtenhofer2021large} and use our self-supervised trained R3D-50 as the backbone for a Faster R-CNN detector. We then extend the 2D RoI features into 3D along the temporal axis, and apply RoIAlign and temporal global average pooling. The RoI features are then max-pooled and fed to a per-class sigmoid classifier. We also use a similar training schedule as~\cite{feichtenhofer2021large}, except we train for only 20 epochs with batch size 64, and use an initial learning rate of 0.1 with 10x step-wise learning rate decay at epochs 5, 10, and 15.

\OURMETHOD{} outperforms the \BASELINE{} baseline on the new task of action detection, on a dataset sourced from cinematic movies rather than Internet videos. \OURMETHOD{} also outperforms CVRL~\cite{qian2021spatiotemporal} despite CVRL being trained for 5x more epochs and using 2x more pretraining frames. As AVA is highly different from the unlabeled K400 videos used for pretraining (while UCF101 and HMDB51 are highly similar to K400), this further supports our hypothesis that improved positive key diversity and added global context through nearest-neighbor sampling improves generalization on video tasks.

\begin{table}[!t]
  \footnotesize
  \renewcommand{\arraystretch}{1.1}
\begin{center}
  \begin{tabular}{c|c|c}
          Method  & Pretrain Data & Top-1 Acc \\
          \hline
          \textcolor{gray}{Supervised} \cite{feichtenhofer2019slowfast} & \textcolor{gray}{K400} & \textcolor{gray}{21.9} \\
          \hline
          CVRL~\cite{qian2021spatiotemporal} & K400 & 16.3 \\
          \BASELINE{}~\cite{feichtenhofer2021large} & K400 & 18.6 \\
          \textbf{\OURMETHOD{}} & K400 & 19.0 \\
          \hline
        \end{tabular}
\end{center}
\caption{\small{\textbf{Action detection on AVA.} We finetune on AVA using a clip size of $8 \times 8$ and report mAP@0.5 IOU.}}
\label{table:ava_eval}
\end{table}

\subsection{Ablations}
We perform a series of ablations to validate our design decisions. Unless otherwise indicated, we pretrain on full unlabeled K400 (240K videos), and the same settings as the main experiments, except for the ablated parameters.

\vspace{-0.2cm}\subsubsection{MLP Heads}
We first ablate the use of separate MLP heads during pretraining. We trained a version of \OURMETHOD{} that shares the MLP head for both intra-video and NN pairs. In this case, the pretraining loss fails to converge and downstream task results are poor. We hypothesize this is due to the different feature spaces learned for intra-video clips vs. inter-video NNs. Thus, we share the backbone but use separate MLP heads during pretraining. No MLP is used for downstream eval.

\vspace{-0.2cm}\subsubsection{Task Generalization of Intra and NN Weights}
In Table \ref{table:nn_weight}, we summarize the results from above sections in which \OURMETHOD{} outperformed \BASELINE{} on every dataset per task and per evaluation protocol. We also ablate our choice of $\lambda_{\textrm{\emph{NN}}}$ by providing results for ($\lambda_{\textrm{\emph{Intra}}}$=0.0, $\lambda_{\textrm{\emph{NN}}}$=1.0), and ($\lambda_{\textrm{\emph{Intra}}}$=1.0, $\lambda_{\textrm{\emph{NN}}}$=1.0). Note that ($\lambda_{\textrm{\emph{Intra}}}$=0.0, $\lambda_{\textrm{\emph{NN}}}$=1.0) corresponds to a pure NN sampling strategy that uses no intra-video pairs, aka a video-analog of NNCLR~\cite{Dwibedi_2021_ICCV}. We summarize the average rank per task for each configuration. The best configuration achieves lowest average rank of 1.2.

Pure NN sampling (video analog of NNCLR~\cite{Dwibedi_2021_ICCV}) is surprisingly competitive with pure intra-video sampling on every task, despite learning zero local semantics during SSL pretraining. However, combining the intra and NN loss ($\lambda_{\textrm{\emph{Intra}}}$=1.0, $\lambda_{\textrm{\emph{NN}}}$=1.0) does not result in as large of a boost as expected. This may suggest that local and global information are complimentary, but may still provide orthogonal directions that are challenging to learn from.

\begin{table}[!t]
  \footnotesize
  \setlength{\tabcolsep}{5pt}
	\begin{center}
		\begin{tabular}{c|ccc|cc}
            & \multicolumn{3}{c|}{UCF Finetune} & \multicolumn{2}{c}{K400 Linear} \\
            \hline
            Method & 1\% & 5\% & 20\% & 1\% & 10\% \\ 
            \hline
            \textcolor{gray}{Supervised (Scratch)} & & & & \textcolor{gray}{3.2} & \textcolor{gray}{39.6} \\
            \hline
            \BASELINE{} \cite{feichtenhofer2021large} & 41.8 & 68.0 & 84.7 & 34.3 & 53.3 \\
            \hline
            \textbf{\OURMETHOD{}} & 44.3 & 68.9 & 85.0 & 34.9 & 54.2 \\
            $\vartriangle$ & \textbf{+2.5} & +0.9 & +0.3 & +0.6 & +0.9 \\
          \end{tabular}
	\end{center}
	\caption{\small{\textbf{Few-shot learning on UCF101 and K400.} Rows indicate different pretrained models on K400. Columns vary the \% of UCF training data used for finetuning and \% of K400 training data used for linear eval.}}
	\label{table:few_shot_learning}
\end{table}

\begin{table}[!t]
  \footnotesize
\begin{center}
  \begin{tabular}{c|c|c|c|c|c}
          Epochs  & $\rho$ & UCF & HMDB & K400 & SSv2 \\
          \hline
          200 & 2 & 92.6 & 65.8 & 65.7 & 53.8 \\
          \hline
          200 & 4 & 93.3 & 67.8 & 66.6 & 54.6 \\
          \hline
          400 & 2 & 93.3 & 68.1 & 67.1 & 54.2 \\
          \hline
        \end{tabular}
\end{center}
\caption{\small{\textbf{More pretraining epochs and NNs.} Data is unlabeled K400.}}
\label{table:more_epochs_nns}
\end{table}

\subsection{Effect of More Epochs and More NNs}
Table~\ref{table:more_epochs_nns} shows that downstream accuracy increases with the number of temporal samples per video and duration of pretraining. Due to limited compute we do not test more combinations, but expect further room for improvement.

\vspace{-0.2cm}\subsection{Few-Shot Learning}
Another way to measure the quality of a self-supervised representation is through data efficiency, or how performance varies with respect to the amount of data available for downstream classification \cite{newell2020useful}. Many works report that self-supervised representations are more data efficient than supervised representations \cite{qian2021spatiotemporal,xiao2021modist,chen2021shot,he2020momentum}.

We first compare against \BASELINE{} on UCF101 using the finetune protocol when training data is limited to 1\%, 5\%, and 20\%, and the evaluation set remains the same. We observe that \OURMETHOD{} is more data efficient across all three subsets. We then compare against \BASELINE{} on K400 using the linear evaluation protocol when training data is limited to 1\% and 10\%, and the evaluation set remains the same. We observe similar improvements across both subsets for \OURMETHOD{}. For both UCF and K400, the delta between \OURMETHOD{} and \BASELINE{} is largest for the smallest training set of 1\% data, indicating that inter-video nearest-neighbors are particularly helpful for generalizing to few-shot settings. We also significantly outperform the supervised baseline in which R3D-50 is initialized from scratch. See Table \ref{table:few_shot_learning}.

\begin{table*}[!t]
    \small
	\begin{center}
		\begin{tabular}{cc|c|c|c|c|c|c}
      \multicolumn{2}{c|}{} & \multicolumn{4}{c|}{Action Recognition} & Action Detection \\
      \cline{3-7}
      \multicolumn{2}{c|}{Model} & \multicolumn{3}{c|}{Finetune} & Linear & Finetune & \textbf{Avg.}  \\
      \cline{3-7}
      $\lambda_{\textrm{\emph{Intra}}}$ & $\lambda_{\textrm{\emph{NN}}}$ & UCF & HMDB & SSv2 & K400 & AVA & \textbf{Rank} \\
      \hline
      1.0 & 0.0 & 91.1 (\textcolor{blue}{\#3}) & 65.3 (\textcolor{blue}{\#3}) & 53.6 (\textcolor{blue}{\#2}) & 65.4 (\textcolor{blue}{\#2}) & 18.6 (\textcolor{blue}{\#2}) & 2.4 \\
      1.0 & 1.0 & \textbf{92.6} (\textcolor{blue}{\#1}) & 65.8 (\textcolor{blue}{\#2}) & \textbf{53.8} (\textcolor{blue}{\#1}) & \textbf{65.7} (\textcolor{blue}{\#1}) & \textbf{19.0} (\textcolor{blue}{\#1}) & 1.2 \\
      0.0 & 1.0 & 91.2 (\textcolor{blue}{\#2}) & \textbf{66.2} (\textcolor{blue}{\#1}) & 53.2 (\textcolor{blue}{\#3}) & 63.7 (\textcolor{blue}{\#3}) & 18.4 (\textcolor{blue}{\#3}) & 2.4 \\
      \hline
    \end{tabular}
	\end{center}
	\caption{\small{\textbf{Do NNs lead to better generalization?} The first row corresponds to the \BASELINE{} baseline and second row corresponds to \OURMETHOD{}. All models are pretrained on full K400 for 200 epochs. Downstream eval uses clip size of $8 \times 8$. $\lambda_{\textrm{\emph{Intra}}}$=0.0 means no intra-video positives are used (NNCLR for video). We denote rank in blue parenthesis (where 1st = best) on each task to show the generalization of each model.}}
	\label{table:nn_weight}
\end{table*}

\begin{table*}[!tb]
  \renewcommand{\arraystretch}{1.1}
  \footnotesize
\begin{center}
  \begin{tabular}{ccc|c|c|c|c|c|c|c|c}
         \multicolumn{3}{c|}{} &\multicolumn{4}{c|}{UCF} & \multicolumn{4}{c}{HMDB} \\ 
          \hline
          Method & Network & Pretrain & R@1 & R@5 & R@10 & R@20 & R@1 & R@5 & R@10 & R@20 \\
          \hline
          SpeedNet \cite{benaim2020speednet} & S3D-G & K400 & 13.0 & 28.1 & 37.5 & 49.5 & & & \\
          GDT \cite{patrick2020multi} & R(2+1)D & K400 & 57.4 & 73.4 & 80.8 & 88.1 & 25.4 & 51.4 & 63.9 & 75.0 \\
          VCLR \cite{kuang2021video} & R2D-50 & K400 & 70.6  & 80.1 & 86.3 & 90.7 & 35.2 & 58.4 & 68.8 & 79.8 \\
          \hline
          \BASELINE{} \cite{feichtenhofer2021large} & R3D-50 & K400 & 73.2 & 87.0 & 91.8 & 95.5 & 36.3 & 61.9 & 72.0 & 82.5 \\
          \hline
          \textbf{\OURMETHOD{}} & R3D-50 & K400 & 74.2 & 87.6 & 92.1 & 95.1 & 37.6 & 62.2 & 72.9 & 82.5 \\
          \hline
        \end{tabular}
\end{center}
\caption{\small{\textbf{Zero-shot video retrieval on UCF101 and HMDB.} We do not compare against two-stream methods. This work only uses RGB.}}
\label{table:retrieval}
\end{table*}

\subsection{Zero-Shot Video Retrieval}
We also evaluate on video retrieval where the extracted features are directly used to find the nearest-neighbors, so there is no further training. Following common practice \cite{han2020self,xu2019self}, we use test-set videos to search the $k$ nearest-neighboring video samples from the train set. We evaluate using Recall at $k$ (R@$K$), which means that the retrieval is correct if any of the top $k$ nearest-neighbors are from the same class as the query. See Table \ref{table:retrieval}.

\OURMETHOD{} outperforms \BASELINE{} for all but one recall threshold on UCF and all recall thresholds on HMDB. This indicates that even without any downstream training, \OURMETHOD{} is better able to push similar videos of a different downstream dataset closer in the embedding space.

\section{Discussion}
\label{sec:discussion}
\subsection{Co-Occurrence of Semantic Classes During Pretraining}
Although labels are not used during self-supervised pretraining, we analyze the probability that a video in the negative queue belongs to the same class as the query video, to understand why the nearest-neighbor objective is beneficial. Assume the queue samples are uniformly sampled and that each class is balanced. Let the pretraining dataset have $K$ balanced classes, and the queue have $Q$ uniformly sampled samples where $Q$ is smaller than the size of the dataset. Then the probability of the above event is 1 - $[(K-1)/K] ^ Q$, which is well over 0.9 for $K$=400 (number of classes in Kinetics-400), $Q$=1024. Note that this calculation also applies to approaches that sample negatives from the mini-batch; let $Q$ be the mini-batch size. CVRL \cite{qian2021spatiotemporal} uses a mini-batch size of 1024 during pretraining. Thus, it is extremely likely that videos belonging to same class as the query are pushed away in the embedding space as negatives in works like \cite{he2020momentum,chen2020simple,qian2021spatiotemporal,feichtenhofer2021large}. Our work does not address this issue by removing poor choices of negatives from the negative set, but rather leverages those similar videos as additional positive keys for a second loss term via the NN sampling strategy, thus providing additional sources of similarity to learn from that would otherwise be ignored.

Additionally, by dynamically computing the positive key using the learned representation space and sampling videos globally, we allow the model to continually evolve its notion of semantic similarity; the quality of the chosen NNs improves as the model learns as demonstrated by~Fig. \ref{fig:nearest_neighbor_evolution}. With intra-video positive pair sampling, the learned representation is not used to choose the positive pairs --- two clips are simply randomly sampled from within a single video.

\subsection{Limitations and Future Work}
While the focus of our work was on improving the diversity of positive keys and balancing global with local notions of similarity, our method can be improved by reducing false negatives similar to works such as~\cite{chuang2020debiased}. Our method could also be improved by leveraging audio-video correspondence~\cite{morgado2021audio} and exploring more nuanced ways to combine the intra and inter-video positives.

%% file: main/6_conclusion.tex
\section{Conclusion}
\label{sec:conclusion}
We presented \OURMETHOD{}, which addresses limitations of existing self-supervised contrastive learning works for video \cite{qian2021spatiotemporal,feichtenhofer2021large} that sample only intra-video clips as positives. By leveraging nearest-neighboring samples from a global neighborhood as positives, we expand the positive keyset beyond individual video boundaries, thus improving the diversity of positive keys and blending global with local context. We then demonstrated that our method improves performance over a wide range of video-related tasks, compared to our intra-video contrastive learning baseline. Our method is simple, effective, and directly plugs into existing contrastive frameworks. \OURMETHOD{} scales to large datasets, unlike methods that require clustering or offline positive sets. We believe that our method has the potential to excel on large unstructured datasets, where self-supervised learning can reach its full potential for video understanding.


%% file: supplement.tex
\section{Other Experiments}
\subsection{Effect of Weight Initialization}
One potential bottleneck to performance is whether random initialization makes it difficult to pick quality NNs during self-supervised pretraining. To explore this, we tried initializing the model with weights with \BASELINE{} pretrained for 200 epochs ($\lambda_{\textrm{\emph{Intra}}}$ = 1.0, $\lambda_{\textrm{\emph{NN}}}$ = 0.0). We then train for 200 more epochs with ($\lambda_{\textrm{\emph{Intra}}}$ = 1.0, $\lambda_{\textrm{\emph{NN}}}$ = 1.0). Note that the effective number of epochs is then 400. Comparing random init at 400 epochs to the pretrained weight init at 200 epochs, there is a not a significant improvement. This indicates that figuring out better ways of blending inter-video NNs and intra-video pairs is a topic for future research.

\begin{table}[!ht]
  \footnotesize
\begin{center}
  \begin{tabular}{c|c|c|c|c|c}
          Weight Init & Pretrain Epochs & UCF & HMDB & K400 & SSv2 \\
          \hline
          Random & 200 & 92.6 & 65.8 & 65.7 & 53.8 \\
          \hline
          Random & 400 & 93.3 & 68.1 & 67.1 & 52.2 \\
          \hline
          \BASELINE{} & 200 (+200) & 92.9 & 67.7 & 66.8 & 54.3 \\
          \hline
        \end{tabular}
\end{center}
\caption{\small{\OURMETHOD{} from random initialization vs. \BASELINE{} weight initialization. Both are trained for 200 epochs with $\rho$=200 on unlabeled K400.}}
\label{table:more_epochs_nns}
\end{table}

\section{Additional Implementation Details}

\subsection{Downstream Evaluation Details}
We mostly follow \cite{feichtenhofer2021large} for downstream eval.

\subsubsection{UCF and HMDB Action Recognition}
We randomly sample a $8 \times 8$ clip and finetune with the same augmentation as during pretraining, but without Gaussian blur. We train for 200 epochs using a batch size of 64 and a base learning rate of 0.005 (8 GPUs) with half period cosine decay. Weight decay is 1e-4 and dropout is 0.5. At test-time, we use standard 10x3 crop evaluation~\cite{feichtenhofer2019slowfast}.

\subsubsection{Kinetics-400 Action Recognition}
\label{supp:eval:k400}
We train a linear classifier on top of the frozen backbone and do not $\ell_{2}$ normalize the embeddings. We randomly sample a $8 \times 8$ clip, randomly resize the shortest side to [256, 320], apply random cropping to $224 \times 224$, and random horizontal flip. We train for 60 epochs using a batch size of 512 and a base learning rate of 0.5 (8 GPUs) with half period cosine decay. Weight decay is 0.0 and dropout is 0.0. At test-time, we use standard 10x3 crop evaluation~\cite{feichtenhofer2019slowfast}.

\subsubsection{Something-Something v2 Action Recognition}
We randomly sample a $8 \times 8$ clip following the same segment-based frame sampling procedure as~\cite{feichtenhofer2021large}. We apply the same augmentations as for K400, but remove flipping because directionality constitutes different action classes in SSv2. We train for 22 epochs using a batch size of 64 and base learning rate of 0.12 (8 GPUs) with 10x step-wise learning rate decay at epochs 14 and 18. Weight decay is $10^{-6}$ and dropout is 0.5. At test-time, we use a single centered $8 \times 8$ clip for evaluation.

\subsubsection{AVA Action Detection}
We randomly sample a $8 \times 8$ clip and finetune with the same augmentations as for K400.  We train for 20 epochs with linear warm-up for first 5 epochs, using a batch size of 64 and base learning rate of 0.1 with step-wise 10x decay at epochs 5, 10, and 15. Weight decay is $10^{-7}$ and dropout is 0.5.

We follow the same region proposal extraction procedure as~\cite{feichtenhofer2021large}. We use an off-the-shelf person detector that is not jointly trained with the action detection models. Specifically we use the predictions provided by the Faster R-CNN with ResNeXt-101-FPN backbone that is provided by the SlowFast~\cite{feichtenhofer2019slowfast} codebase. At test-time, we use a single centered $8 \times 8$ clip for evaluation.

\subsection{Encoder Architecture}
Our default encoder is a R3D-50 Slow model \cite{feichtenhofer2019slowfast}, with a temporal dimension of size $T$=8 and sample rate $\tau$=8. Architecture diagram is given in Table \ref{table:encoder_architecture}.
\begin{table}[!t]
    \begin{center}
    \begin{tabular}{c|c|c}
        stage & kernels & output sizes $T \times S^2$ \\
        \hline
        raw clip & - & $T\tau \times 224^2$ \\
        \hline
        data layer & stride $\tau$, $1^2$ & $T \times 224^2$ \\
        \hline
        \multirow{2}{*}{$\textrm{conv}_1$} & $1 \times 7^2$, 64 & \multirow{2}{*}{$T \times 112^2$} \\
        & stride 1, $2^2$ & \\
        \hline
        \multirow{2}{*}{$\textrm{pool}_1$} & $1 \times 3^2$ max & \multirow{2}{*}{$T \times 56^2$} \\
        & stride 1, $2^2$ & \\
        \hline
        \multirow{4}{*}{$\textrm{res}_2$} & ($\times$ 3) & \multirow{4}{*}{$T \times 56^2$} \\
        & $1 \times 1^2$, 64 & \\
        & $1 \times 3^2$, 64 & \\
        & $1 \times 1^2$, 256 & \\
        \hline
        \multirow{4}{*}{$\textrm{res}_3$} & ($\times$ 4) & \multirow{4}{*}{$T \times 28^2$} \\
        & $1 \times 1^2$, 128 & \\
        & $1 \times 3^2$, 128 & \\
        & $1 \times 1^2$, 512 & \\
        \hline
        \multirow{4}{*}{$\textrm{res}_4$} & ($\times$ 6) & \multirow{4}{*}{$T \times 14^2$} \\
        & $3 \times 1^2$, 256 & \\
        & $1 \times 3^2$, 256 & \\
        & $1 \times 1^2$, 1024 & \\
        \hline
        \multirow{4}{*}{$\textrm{res}_5$} & ($\times$ 3) & \multirow{4}{*}{$T \times 7^2$} \\
        & $3 \times 1^2$, 512 & \\
        & $1 \times 3^2$, 512 & \\
        & $1 \times 1^2$, 2048 & \\
        \hline
        $\textrm{pool}_5$ & global average pool & $1 \times 1^2$ \\
    \end{tabular}
    \end{center}
    \caption{We use the R3D-50 Slow pathway \cite{feichtenhofer2019slowfast} as our 3D encoder for self-supervised learning. The dimensions of kernels are given as \{$T \times S^2, C$\} for temporal, spatial, and channel sizes. Strides are given as \{temporal stride, $\textrm{spatial stride}^2$\}. Non-degenerate temporal filters are underlined. Temporal pooling is only performed at the last layer to collapse the space and time dimensions. By default, we sample clips at $T \times \tau$ = 8 x 8.}
    \label{table:encoder_architecture}
\end{table}

\section{Additional Qualitative Results}
We provide additional qualitative examples to help visualize what the model is learning.

\subsection{Nearest-Neighbor Evolution During Pretraining}
\begin{figure*}[!t]
	\centering

	\begin{subfigure}[t]{0.48\textwidth}
		\centering
		\includegraphics[width=0.95\linewidth]{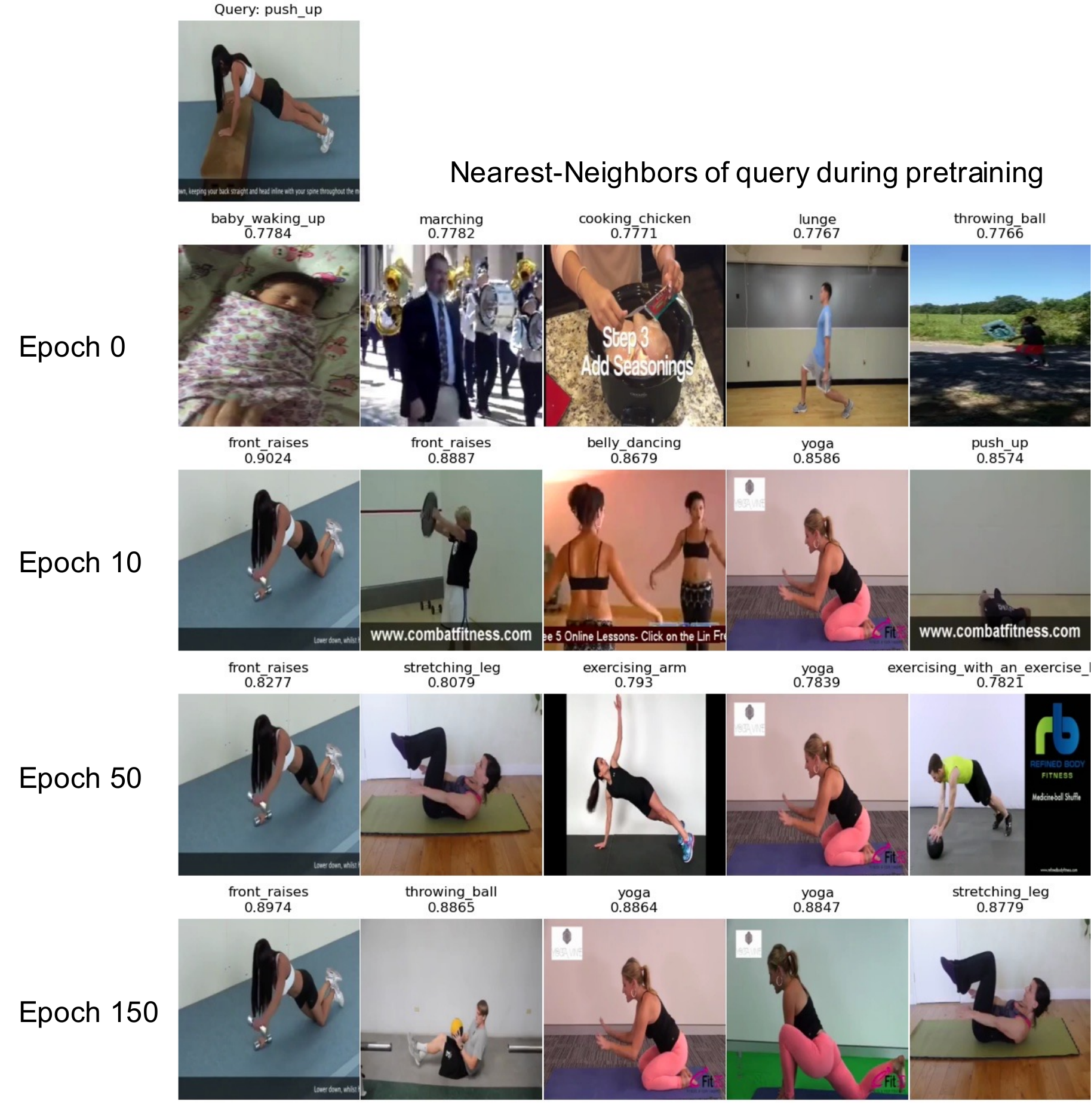}
        \caption{}
		\label{fig:nn_evolution_pushup}
	\end{subfigure}
	~
	\begin{subfigure}[t]{0.48\textwidth}
		\centering
		\includegraphics[width=0.95\linewidth]{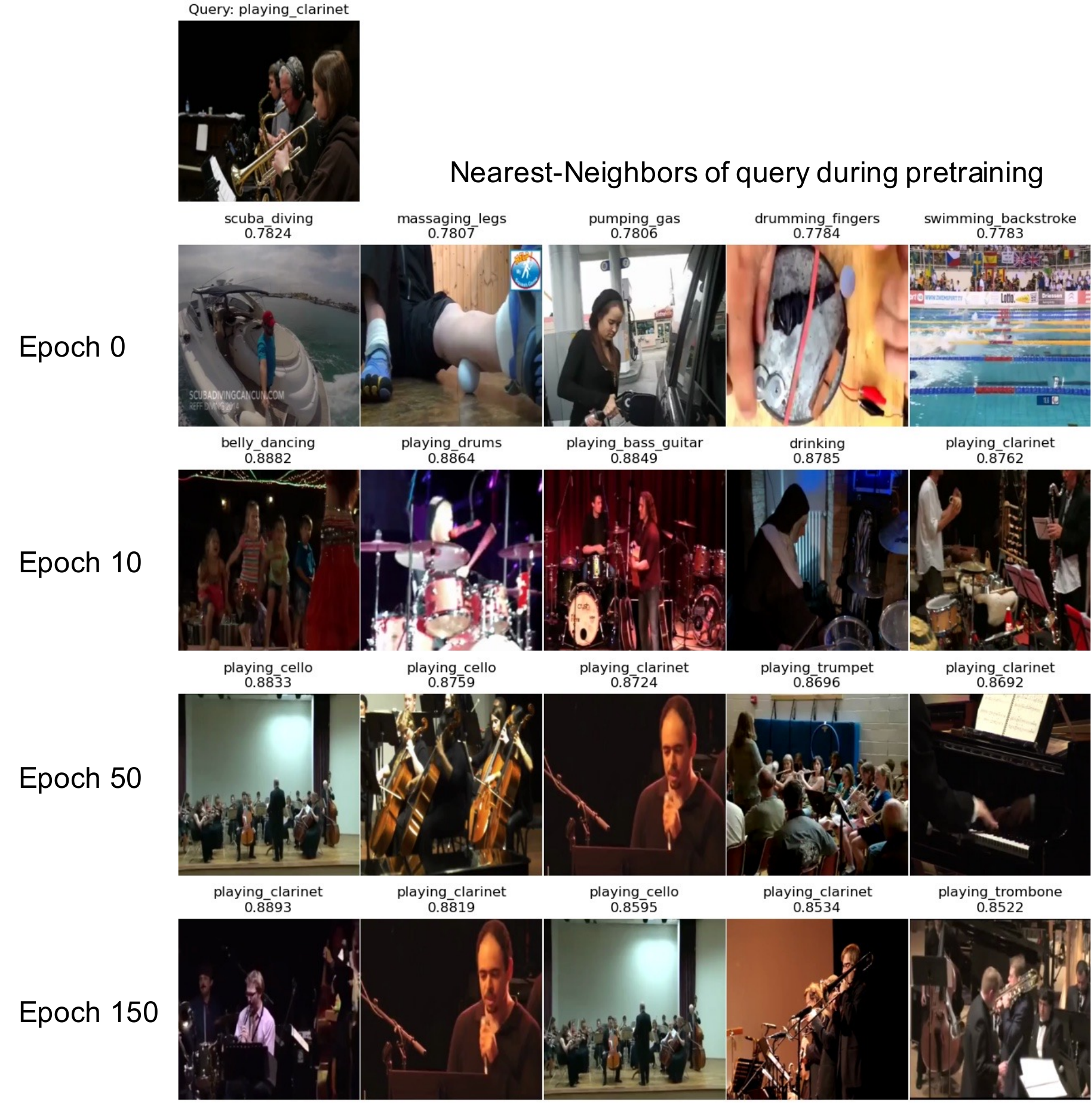}
        \caption{}
		\label{fig:nn_evolution_clarinet}
	\end{subfigure}
    ~
	\begin{subfigure}[t]{0.48\textwidth}
		\centering
		\includegraphics[width=0.95\linewidth]{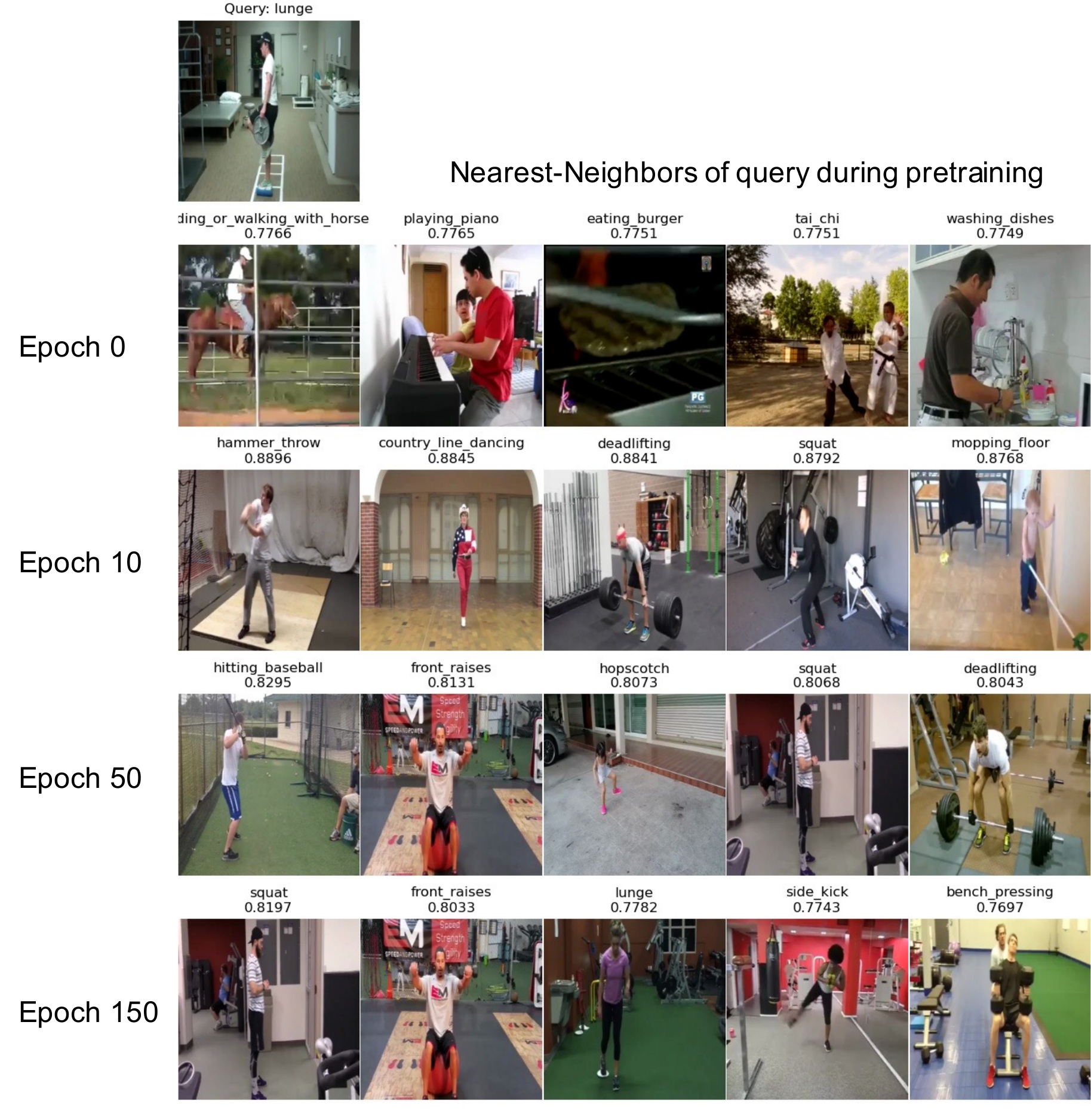}
        \caption{}
		\label{fig:nn_evolution_lunge}
	\end{subfigure}
    ~
	\begin{subfigure}[t]{0.48\textwidth}
		\centering
		\includegraphics[width=0.95\linewidth]{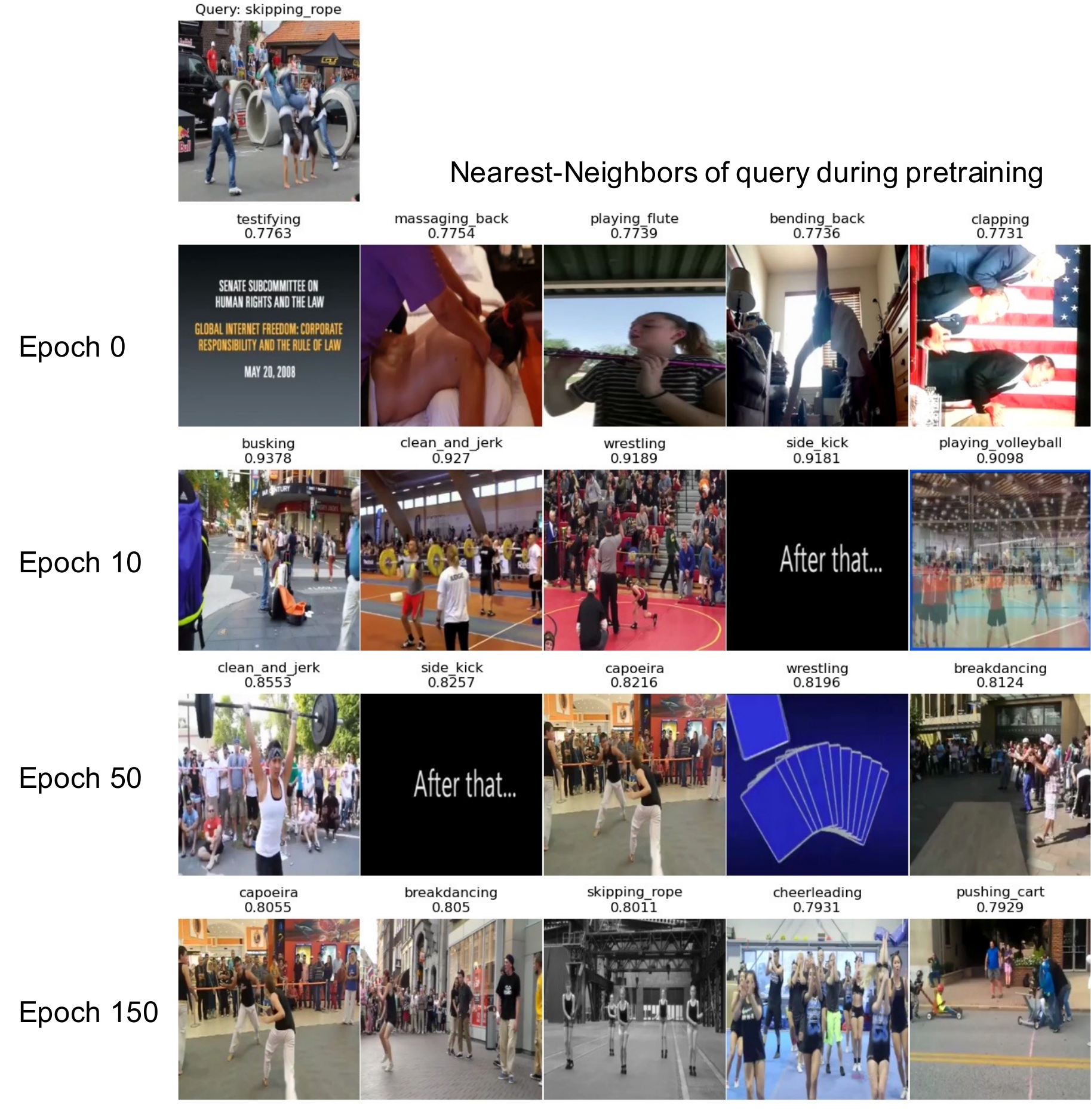}
        \caption{}
		\label{fig:nn_evolution_skiprope}
	\end{subfigure}

	\caption{Evolution of inter-video nearest-neighbor positives during pretraining, starting from random initialization. Videos are sampled across the dataset using similarity measured by the learned feature space, which improves during pretraining. Top row is the query video while other rows indicate different epochs of pretraining.}
	\label{fig:nn_evolution}
\end{figure*}

In Figure \ref{fig:nn_evolution}, we show how the nearest-neighbors vary over the course of pre-training, starting from random initialization. We observe that as training progresses, the nearest-neighbors become more semantically similar, which complements Figure 3b) of the main paper.

\subsection{Downstream Zero-Shot Retrieval}
As mentioned in $\S4.8$ of the main paper, extracted features from the test set are directly used to retrieve nearest-neighboring training set samples. No further training is done. While our retrieved results may sometimes belong to a different class as the query, they are still semantically similar to the query.

\section{Self-Supervised Pre-Training Visualizations}
To provide a visualization of the semantic feature clusters during the course of unsupervised pretraining on K400, we plotted t-SNE feature visualization on UCF101. The model is not finetuned on UCF. With more pre-training, the features organize into semantically related clusters. For example, ``ApplyEyeMakeup'' and ``ApplyLipstick'' cluster together because they are acts of applying cosmetics.

\begin{figure*}[!tb]
	\centering

	\begin{subfigure}[t]{0.48\textwidth}
		\centering
		\includegraphics[width=0.95\linewidth]{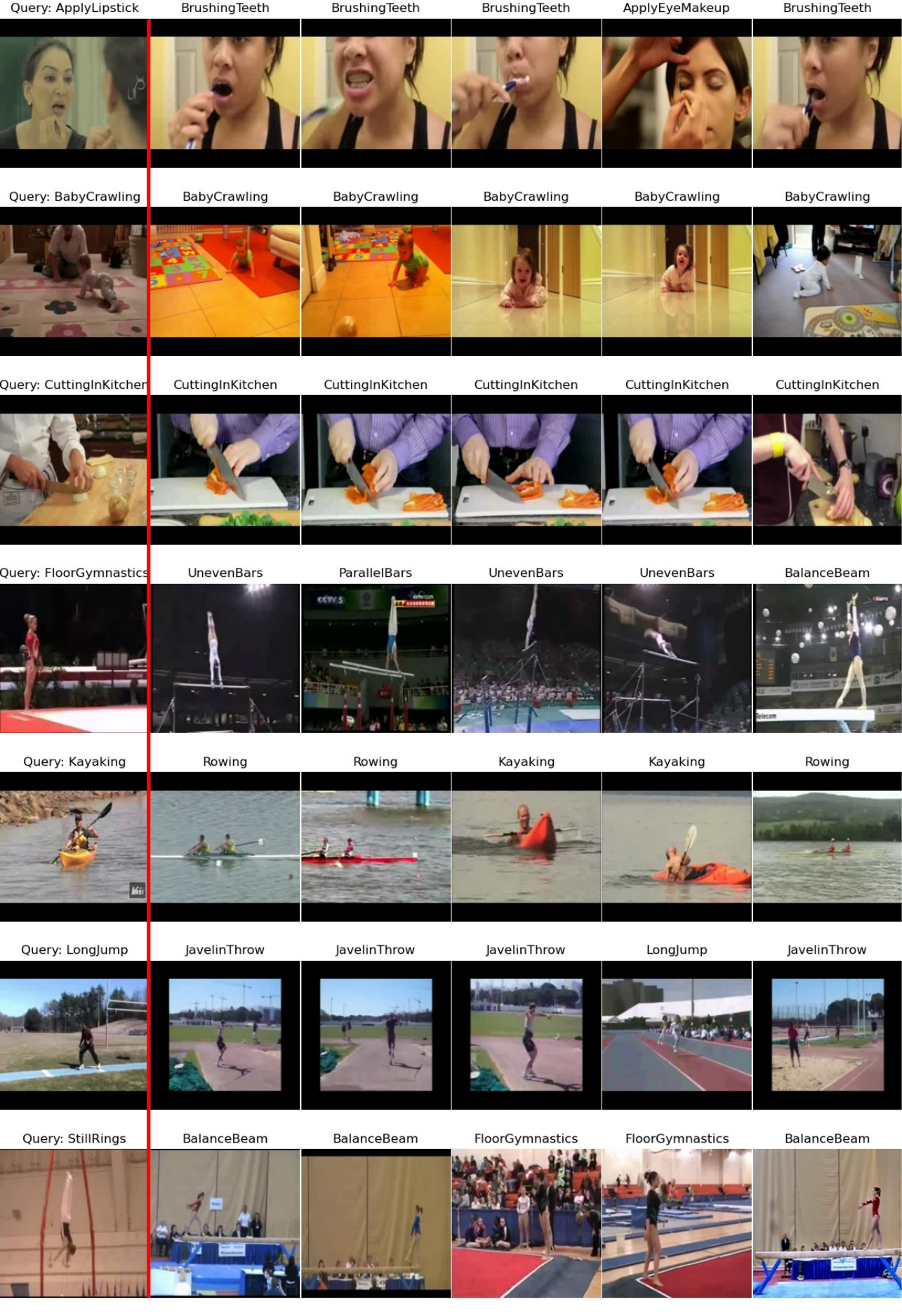}
        \caption{UCF101 retrieval results.}
		\label{fig:retrieval:ucf}
	\end{subfigure}
	~
	\begin{subfigure}[t]{0.48\textwidth}
		\centering
		\includegraphics[width=0.95\linewidth]{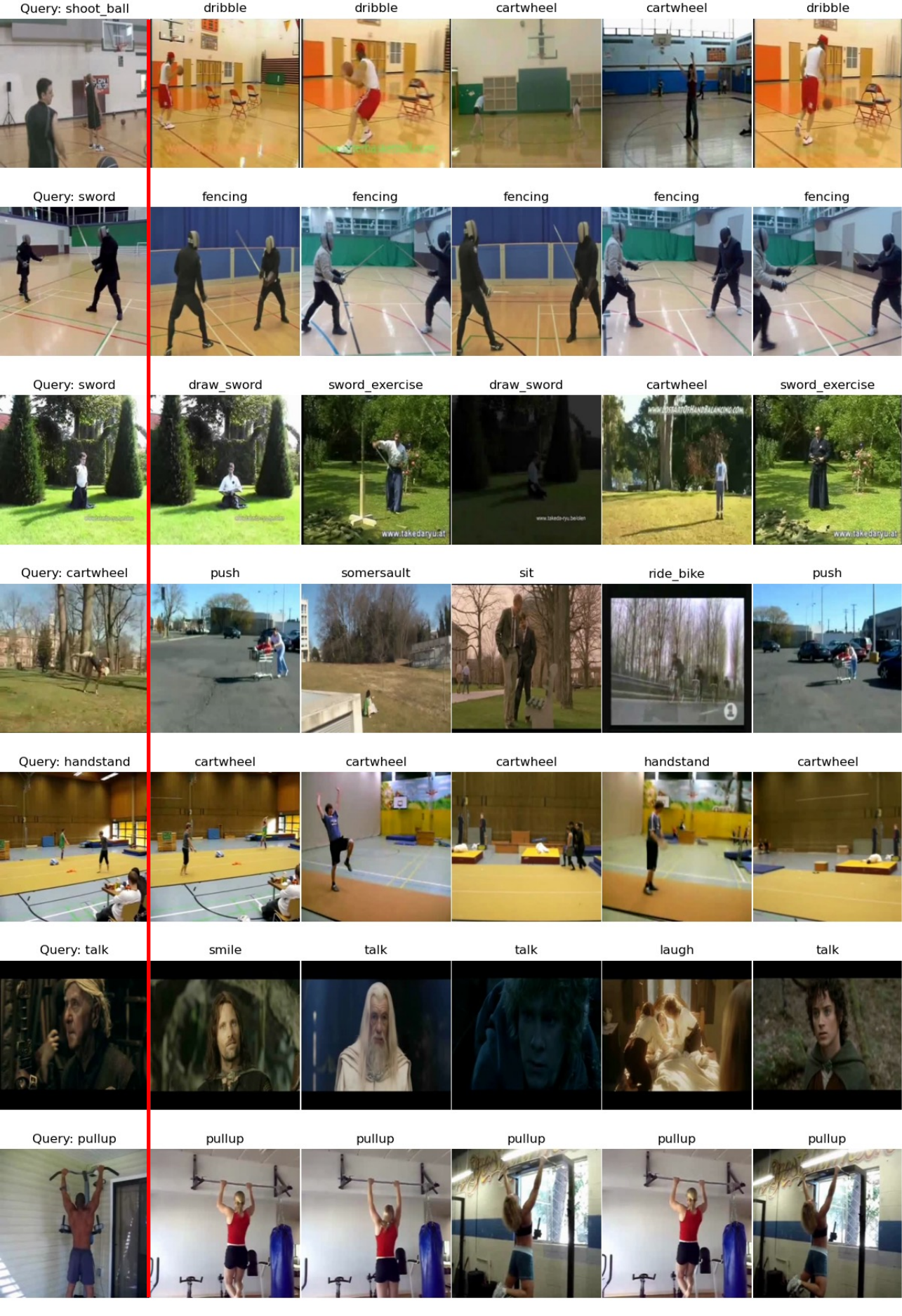}
        \caption{HMDB51 retrieval results.}
		\label{fig:retrieval:hmdb}
	\end{subfigure}

	\caption{Video retrieval results for \OURMETHOD{} on a) UCF101 and b) HMDB51. The left-most column shows the query video while the remaining columns show the top-5 nearest neighbors.}
	\label{fig:retrieval_visualization}
\end{figure*}

\begin{figure*}[!th]
	\centering
        \includegraphics[width=0.95\linewidth]{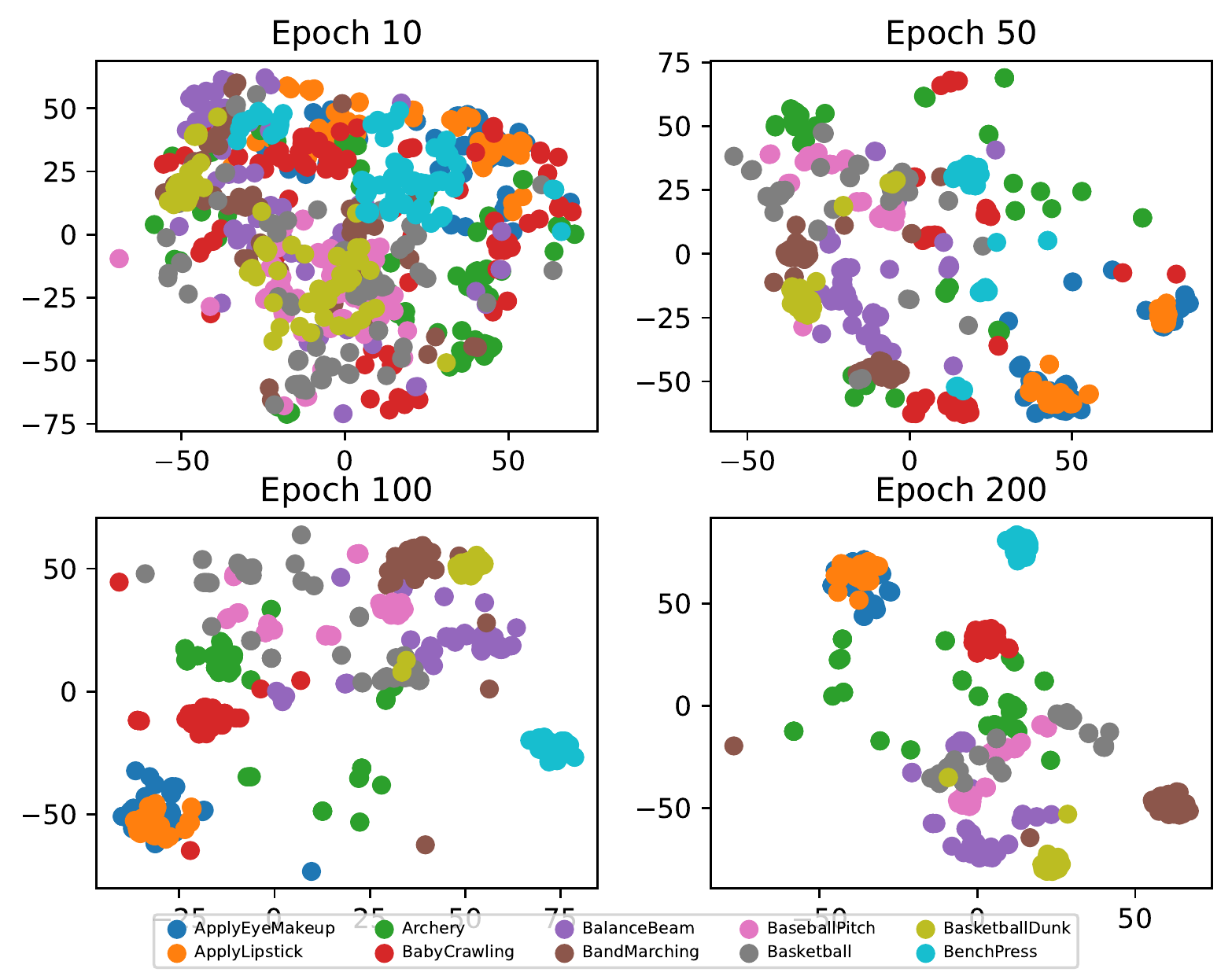}
        \caption{t-SNE feature visualization over epochs of pre-training.}
        \label{fig:tsne_visualization}
\end{figure*}

\clearpage
\section{Pseudocode}
In Algorithm \ref{algorithm_momentum}, we present the pretraining pseudocode for \OURMETHOD{}. Note that our approach does not require the usage of momentum encoders. Algorithm \ref{algorithm_nonmomentum} shows the pretraining pseudocode \textbf{without} momentum encoders.

\begin{algorithm*}
\tiny
\caption{Pseudocode for \OURMETHOD{} pretraining with momentum encoder.}
\begin{lstlisting}[language=Python]
# f_q: online encoder
# g_q_intra, g_q_nn: online projection MLP for intra loss and NN loss
# f_k: momentum encoder
# g_k_intra, g_k_nn: momentum projection MLP for intra loss and NN loss
# m: momentum update coeff.
# intra_queue, nn_queue: queue for intra loss (KxC) and NN loss (KxC)
# intra_lambda, nn_lambda: intra loss and NN loss weights

# load minibatch with N samples. Two clips x1, x2 are sampled from each video.
for (x1, x2) in loader:
    x1, x2 = aug(x1), aug(x2) # random augmentation
    q1, q2 = f_q(x1), f_q(x2) # forward pass to online encoder (NxC)
    # Apply online projection MLPs to view 1
    q1_intra, q1_nn = g_q_intra(q1), g_q_nn(q1)
    # Apply online projection MLPs to view 2
    q2_intra, q2_nn = g_q_intra(q2), g_q_nn(q2)
    k1, k2 = f_k(x1), f_k(x2) # forward pass to momentum encoder (NxC)
    # Apply momentum projection MLPs to view 1
    k1_intra, k1_nn = g_k_intra(k1), g_k_nn(k1)
    # Apply momentum projection MLPs to view 2
    k2_intra, k2_nn = g_k_intra(k2), g_k_nn(k2)
    #  No gradients to keys
    k1_intra, k1_nn, k2_intra, k2_nn = detach(k1_intra, k1_nn, k2_intra, k2_nn)
   
    # Symmetric loss
    total_loss = L(q1_intra, q1_nn, k2_intra, k2_nn) + 
                L(q2_intra, q2_nn, k1_intra, k1_nn)
    total_loss.backward() # backprop
    update(f_q.params), update(g_q_intra.params), update(g_q_nn.params) # SGD update
    # update momentum encoder
    f_k.params = m * f_k.params + (1-m) * f_q.params
    g_k_intra.params = m * g_k_intra.params + (1-m) * g_q_intra.params
    g_k_nn.params = m * g_k_nn.params + (1-m) * g_q_nn.params
    # update queues
    update_queue(intra_queue, k1_intra)
    update_queue(nn_queue, k1_nn)

def L(q_intra, q_nn, k_intra, k_nn, intra_lambda=1.0, nn_lambda=0.2):
    nn_idx, nn_loss = L_NN(q_nn, k_nn, nn_queue)
    intra_loss = L_Intra(q_intra, k_intra, nn_idx, intra_queue)
    return intra_lambda * intra_loss + nn_lambda * nn_loss

def L_NN(q, k, nn_queue):
    q, k = normalize(q, dim=1), normalize(k, dim=1) # L2 normalize the embeddings
    # find the location of the nearest-neighbor of k1_nn and k2_nn in nn_queue
    nn_idx = NN(k, nn_queue) # N
    nn_logits = mm(q.view(N,C), nn_queue.view(C,K)) # nn_logits: NxK
    nn_loss = cross_entropy(nn_logits, nn_idx)
    return nn_idx, nn_loss

def L_Intra(q, k, nn_idx, intra_queue):
    q, k = normalize(q, dim=1), normalize(k, dim=1) # L2 normalize the embeddings
    # maskout nearest-neighbor positives from intra_queue
    intra_mask = ones(intra_queue.shape[0]) # mask: Kx1 
    intra_mask[nn_idx] = False 
    masked_intra_queue = intra_queue[intra_mask]
    # intra positive logits: Nx1
    intra_logits_positive = bmm(q.view(N,1,C), k.view(N,C,1))
    # intra negative logits: NxK 
    intra_logits_negative = mm(q.view(N,C), masked_intra_queue.view(C,K)) 
    # intra logits: N x(1+K)
    intra_logits = cat([intra_logits_positive, intra_logits_negative], dim=1) 
    intra_label = zeros(N)
    intra_loss = cross_entropy(intra_logits, intra_label)
    return intra_loss

# find the location of nearest-neighbor embedding in the queue
def NN(z,Q):
    z = normalize(z, dim=1), Q = normalize(Q, dim=1) # L2 normalize
    # z: input embedding minibatch, NxC
    nn_idx = (z@Q.T).argmax(dim=1) # NN location: Nx1
    return nn_idx
\end{lstlisting}
    \label{algorithm_momentum}
\end{algorithm*}

\begin{algorithm*}
\tiny
\caption{Pseudocode for \OURMETHOD{} pretraining \textbf{without} momentum encoder.}
\begin{lstlisting}[language=Python]
# f: encoder
# g_intra, g_nn: online projection MLP for intra loss and NN loss
# queue: queue for NN loss (KxC)
# intra_lambda, nn_lambda: intra loss and NN loss weights

# load minibatch with N samples. Two clips x1, x2 are sampled from each video.
for (x1, x2) in loader:
    x1, x2 = aug(x1), aug(x2) # random augmentation
    q1, q2 = f(x1), f(x2) # forward pass to online encoder (NxC)
    q1_intra, q1_nn = g_intra(q1), g_nn(q1) # Apply online projection MLPs to view 1
    q2_intra, q2_nn = g_intra(q2), g_nn(q2) # Apply online projection MLPs to view 2
   
    # Symmetric loss
    total_loss = intra_lambda * L(q1_intra, q2_intra) + \
        nn_lambda * (L(q1_nn, NN(q2_nn,queue)) + L(q2_nn, NN(q1_nn,queue)))

    total_loss.backward() # backprop
    update(f.params), update(g_intra.params), update(g_nn.params) # SGD update

    # update queues
    update_queue(nn_queue, q1_nn)

def L(q, k,temperature=0.1):
    q, k = normalize(q, dim=1), normalize(k, dim=1) # L2 normalize the embeddings
    logits = q@k.T # nn_logits: NxN
    logits /= temperature # apply contrastive loss temperature
    logits_idx = range(0,N)
    contrastive_loss = cross_entropy(logits, logits_idx)
    return contrastive_loss

# return nearest-neighbor embedding in the queue
def NN(z,Q):
    z = normalize(z, dim=1), Q = normalize(Q, dim=1) # L2 normalize
    # z: input embedding minibatch, NxC
    nn_idx = (z@Q.T).argmax(dim=1) # NN location: Nx1
    return Q[nn_idx]
\end{lstlisting}
    \label{algorithm_nonmomentum}
\end{algorithm*}

%% file: main.bbl
\begin{thebibliography}{10}\itemsep=-1pt

\bibitem{behrmann2021long}
Nadine Behrmann, Mohsen Fayyaz, Juergen Gall, and Mehdi Noroozi.
\newblock Long short view feature decomposition via contrastive video
  representation learning.
\newblock In {\em Proceedings of the IEEE/CVF International Conference on
  Computer Vision}, pages 9244--9253, 2021.

\bibitem{benaim2020speednet}
Sagie Benaim, Ariel Ephrat, Oran Lang, Inbar Mosseri, William~T Freeman,
  Michael Rubinstein, Michal Irani, and Tali Dekel.
\newblock Speednet: Learning the speediness in videos.
\newblock In {\em Proceedings of the IEEE/CVF Conference on Computer Vision and
  Pattern Recognition}, pages 9922--9931, 2020.

\bibitem{brown2020language}
Tom~B Brown, Benjamin Mann, Nick Ryder, Melanie Subbiah, Jared Kaplan, Prafulla
  Dhariwal, Arvind Neelakantan, Pranav Shyam, Girish Sastry, Amanda Askell,
  et~al.
\newblock Language models are few-shot learners.
\newblock {\em arXiv preprint arXiv:2005.14165}, 2020.

\bibitem{caron2020unsupervised}
Mathilde Caron, Ishan Misra, Julien Mairal, Priya Goyal, Piotr Bojanowski, and
  Armand Joulin.
\newblock Unsupervised learning of visual features by contrasting cluster
  assignments.
\newblock In {\em Thirty-fourth Conference on Neural Information Processing
  Systems (NeurIPS)}, 2020.

\bibitem{chen2021multimodal}
Brian Chen, Andrew Rouditchenko, Kevin Duarte, Hilde Kuehne, Samuel Thomas,
  Angie Boggust, Rameswar Panda, Brian Kingsbury, Rogerio Feris, David Harwath,
  et~al.
\newblock Multimodal clustering networks for self-supervised learning from
  unlabeled videos.
\newblock {\em arXiv preprint arXiv:2104.12671}, 2021.

\bibitem{chen2021shot}
Shixing Chen, Xiaohan Nie, David Fan, Dongqing Zhang, Vimal Bhat, and Raffay
  Hamid.
\newblock Shot contrastive self-supervised learning for scene boundary
  detection.
\newblock In {\em Proceedings of the IEEE/CVF Conference on Computer Vision and
  Pattern Recognition}, pages 9796--9805, 2021.

\bibitem{chen2020simple}
Ting Chen, Simon Kornblith, Mohammad Norouzi, and Geoffrey Hinton.
\newblock A simple framework for contrastive learning of visual
  representations.
\newblock In {\em International conference on machine learning}, pages
  1597--1607. PMLR, 2020.

\bibitem{chen2020big}
Ting Chen, Simon Kornblith, Kevin Swersky, Mohammad Norouzi, and Geoffrey
  Hinton.
\newblock Big self-supervised models are strong semi-supervised learners.
\newblock {\em arXiv preprint arXiv:2006.10029}, 2020.

\bibitem{chen2020oasis}
Weifeng Chen, Shengyi Qian, David Fan, Noriyuki Kojima, Max Hamilton, and Jia
  Deng.
\newblock Oasis: A large-scale dataset for single image 3d in the wild.
\newblock In {\em Proceedings of the IEEE/CVF Conference on Computer Vision and
  Pattern Recognition}, pages 679--688, 2020.

\bibitem{chen2020improved}
Xinlei Chen, Haoqi Fan, Ross Girshick, and Kaiming He.
\newblock Improved baselines with momentum contrastive learning, 2020.

\bibitem{chen2021exploring}
Xinlei Chen and Kaiming He.
\newblock Exploring simple siamese representation learning.
\newblock In {\em Proceedings of the IEEE/CVF Conference on Computer Vision and
  Pattern Recognition}, pages 15750--15758, 2021.

\bibitem{chuang2020debiased}
Ching-Yao Chuang, Joshua Robinson, Yen-Chen Lin, Antonio Torralba, and Stefanie
  Jegelka.
\newblock Debiased contrastive learning.
\newblock {\em Advances in neural information processing systems},
  33:8765--8775, 2020.

\bibitem{dave2021tclr}
Ishan Dave, Rohit Gupta, Mamshad~Nayeem Rizve, and Mubarak Shah.
\newblock Tclr: Temporal contrastive learning for video representation.
\newblock {\em arXiv preprint arXiv:2101.07974}, 2021.

\bibitem{deng2009imagenet}
Jia Deng, Wei Dong, Richard Socher, Li-Jia Li, Kai Li, and Li Fei-Fei.
\newblock Imagenet: A large-scale hierarchical image database.
\newblock In {\em Proceedings of the IEEE Conference on Computer Vision and
  Pattern Recognition}, 2009.

\bibitem{devlin2018bert}
Jacob Devlin, Ming-Wei Chang, Kenton Lee, and Kristina Toutanova.
\newblock Bert: Pre-training of deep bidirectional transformers for language
  understanding.
\newblock {\em arXiv preprint arXiv:1810.04805}, 2018.

\bibitem{diba2019dynamonet}
Ali Diba, Vivek Sharma, Luc~Van Gool, and Rainer Stiefelhagen.
\newblock Dynamonet: Dynamic action and motion network.
\newblock In {\em Proceedings of the IEEE/CVF International Conference on
  Computer Vision}, pages 6192--6201, 2019.

\bibitem{doersch2015unsupervised}
Carl Doersch, Abhinav Gupta, and Alexei~A Efros.
\newblock Unsupervised visual representation learning by context prediction.
\newblock In {\em Proceedings of the IEEE international conference on computer
  vision}, pages 1422--1430, 2015.

\bibitem{Dwibedi_2021_ICCV}
Debidatta Dwibedi, Yusuf Aytar, Jonathan Tompson, Pierre Sermanet, and Andrew
  Zisserman.
\newblock With a little help from my friends: Nearest-neighbor contrastive
  learning of visual representations.
\newblock In {\em Proceedings of the IEEE/CVF International Conference on
  Computer Vision (ICCV)}, pages 9588--9597, October 2021.

\bibitem{feichtenhofer2019slowfast}
Christoph Feichtenhofer, Haoqi Fan, Jitendra Malik, and Kaiming He.
\newblock Slowfast networks for video recognition.
\newblock In {\em Proceedings of the IEEE/CVF international conference on
  computer vision}, pages 6202--6211, 2019.

\bibitem{feichtenhofer2021large}
Christoph Feichtenhofer, Haoqi Fan, Bo Xiong, Ross Girshick, and Kaiming He.
\newblock A large-scale study on unsupervised spatiotemporal representation
  learning.
\newblock In {\em Proceedings of the IEEE/CVF Conference on Computer Vision and
  Pattern Recognition}, pages 3299--3309, 2021.

\bibitem{fernando2017self}
Basura Fernando, Hakan Bilen, Efstratios Gavves, and Stephen Gould.
\newblock Self-supervised video representation learning with odd-one-out
  networks.
\newblock In {\em Proceedings of the IEEE conference on computer vision and
  pattern recognition}, pages 3636--3645, 2017.

\bibitem{gidaris2018unsupervised}
Spyros Gidaris, Praveer Singh, and Nikos Komodakis.
\newblock Unsupervised representation learning by predicting image rotations.
\newblock {\em arXiv preprint arXiv:1803.07728}, 2018.

\bibitem{goyal2017something}
Raghav Goyal, Samira Ebrahimi~Kahou, Vincent Michalski, Joanna Materzynska,
  Susanne Westphal, Heuna Kim, Valentin Haenel, Ingo Fruend, Peter Yianilos,
  Moritz Mueller-Freitag, et~al.
\newblock The" something something" video database for learning and evaluating
  visual common sense.
\newblock In {\em Proceedings of the IEEE international conference on computer
  vision}, pages 5842--5850, 2017.

\bibitem{grill2020bootstrap}
Jean-Bastien Grill, Florian Strub, Florent Altch{\'e}, Corentin Tallec, Pierre
  Richemond, Elena Buchatskaya, Carl Doersch, Bernardo Pires, Zhaohan Guo,
  Mohammad Azar, et~al.
\newblock Bootstrap your own latent: A new approach to self-supervised
  learning.
\newblock In {\em Neural Information Processing Systems}, 2020.

\bibitem{gu2018ava}
Chunhui Gu, Chen Sun, David~A Ross, Carl Vondrick, Caroline Pantofaru, Yeqing
  Li, Sudheendra Vijayanarasimhan, George Toderici, Susanna Ricco, Rahul
  Sukthankar, et~al.
\newblock Ava: A video dataset of spatio-temporally localized atomic visual
  actions.
\newblock In {\em Proceedings of the IEEE Conference on Computer Vision and
  Pattern Recognition}, pages 6047--6056, 2018.

\bibitem{han2019video}
Tengda Han, Weidi Xie, and Andrew Zisserman.
\newblock Video representation learning by dense predictive coding.
\newblock In {\em Proceedings of the IEEE/CVF International Conference on
  Computer Vision Workshops}, pages 0--0, 2019.

\bibitem{han2020memory}
Tengda Han, Weidi Xie, and Andrew Zisserman.
\newblock Memory-augmented dense predictive coding for video representation
  learning.
\newblock In {\em Computer Vision--ECCV 2020: 16th European Conference,
  Glasgow, UK, August 23--28, 2020, Proceedings, Part III 16}, pages 312--329.
  Springer, 2020.

\bibitem{han2020self}
Tengda Han, Weidi Xie, and Andrew Zisserman.
\newblock Self-supervised co-training for video representation learning.
\newblock {\em Advances in Neural Information Processing Systems},
  33:5679--5690, 2020.

\bibitem{he2020momentum}
Kaiming He, Haoqi Fan, Yuxin Wu, Saining Xie, and Ross Girshick.
\newblock Momentum contrast for unsupervised visual representation learning.
\newblock In {\em Proceedings of the IEEE/CVF Conference on Computer Vision and
  Pattern Recognition}, pages 9729--9738, 2020.

\bibitem{he2015delving}
Kaiming He, Xiangyu Zhang, Shaoqing Ren, and Jian Sun.
\newblock Delving deep into rectifiers: Surpassing human-level performance on
  imagenet classification.
\newblock In {\em Proceedings of the IEEE international conference on computer
  vision}, pages 1026--1034, 2015.

\bibitem{huang2020movienet}
Qingqiu Huang, Yu Xiong, Anyi Rao, Jiaze Wang, and Dahua Lin.
\newblock Movienet: A holistic dataset for movie understanding.
\newblock In {\em Computer Vision--ECCV 2020: 16th European Conference,
  Glasgow, UK, August 23--28, 2020, Proceedings, Part IV 16}, pages 709--727.
  Springer, 2020.

\bibitem{jabri2020space}
Allan Jabri, Andrew Owens, and Alexei Efros.
\newblock Space-time correspondence as a contrastive random walk.
\newblock {\em Advances in neural information processing systems}, 33, 2020.

\bibitem{jenni2021time}
Simon Jenni and Hailin Jin.
\newblock Time-equivariant contrastive video representation learning.
\newblock In {\em Proceedings of the IEEE/CVF International Conference on
  Computer Vision}, pages 9970--9980, 2021.

\bibitem{jing2018self}
Longlong Jing, Xiaodong Yang, Jingen Liu, and Yingli Tian.
\newblock Self-supervised spatiotemporal feature learning via video rotation
  prediction.
\newblock {\em arXiv preprint arXiv:1811.11387}, 2018.

\bibitem{kay2017kinetics}
Will Kay, Joao Carreira, Karen Simonyan, Brian Zhang, Chloe Hillier, Sudheendra
  Vijayanarasimhan, Fabio Viola, Tim Green, Trevor Back, Paul Natsev, et~al.
\newblock The kinetics human action video dataset.
\newblock {\em arXiv preprint arXiv:1705.06950}, 2017.

\bibitem{kim2019self}
Dahun Kim, Donghyeon Cho, and In~So Kweon.
\newblock Self-supervised video representation learning with space-time cubic
  puzzles.
\newblock In {\em Proceedings of the AAAI Conference on Artificial
  Intelligence}, volume~33, pages 8545--8552, 2019.

\bibitem{kolesnikov2019revisiting}
Alexander Kolesnikov, Xiaohua Zhai, and Lucas Beyer.
\newblock Revisiting self-supervised visual representation learning.
\newblock In {\em Proceedings of the IEEE/CVF conference on computer vision and
  pattern recognition}, pages 1920--1929, 2019.

\bibitem{kuang2021video}
Haofei Kuang, Yi Zhu, Zhi Zhang, Xinyu Li, Joseph Tighe, Soren Schwertfeger,
  Cyrill Stachniss, and Mu Li.
\newblock Video contrastive learning with global context.
\newblock In {\em Proceedings of the IEEE/CVF International Conference on
  Computer Vision}, pages 3195--3204, 2021.

\bibitem{kuehne2011hmdb}
Hildegard Kuehne, Hueihan Jhuang, Est{\'\i}baliz Garrote, Tomaso Poggio, and
  Thomas Serre.
\newblock Hmdb: a large video database for human motion recognition.
\newblock In {\em 2011 International conference on computer vision}, pages
  2556--2563. IEEE, 2011.

\bibitem{lee2017unsupervised}
Hsin-Ying Lee, Jia-Bin Huang, Maneesh Singh, and Ming-Hsuan Yang.
\newblock Unsupervised representation learning by sorting sequences.
\newblock In {\em Proceedings of the IEEE International Conference on Computer
  Vision}, pages 667--676, 2017.

\bibitem{li2020prototypical}
Junnan Li, Pan Zhou, Caiming Xiong, and Steven~CH Hoi.
\newblock Prototypical contrastive learning of unsupervised representations.
\newblock {\em arXiv preprint arXiv:2005.04966}, 2020.

\bibitem{loshchilov2016sgdr}
Ilya Loshchilov and Frank Hutter.
\newblock Sgdr: Stochastic gradient descent with warm restarts.
\newblock {\em arXiv preprint arXiv:1608.03983}, 2016.

\bibitem{luo2017unsupervised}
Zelun Luo, Boya Peng, De-An Huang, Alexandre Alahi, and Li Fei-Fei.
\newblock Unsupervised learning of long-term motion dynamics for videos.
\newblock In {\em Proceedings of the IEEE conference on computer vision and
  pattern recognition}, pages 2203--2212, 2017.

\bibitem{misra2016shuffle}
Ishan Misra, C~Lawrence Zitnick, and Martial Hebert.
\newblock Shuffle and learn: unsupervised learning using temporal order
  verification.
\newblock In {\em European Conference on Computer Vision}, pages 527--544.
  Springer, 2016.

\bibitem{morgado2021audio}
Pedro Morgado, Nuno Vasconcelos, and Ishan Misra.
\newblock Audio-visual instance discrimination with cross-modal agreement.
\newblock In {\em Proceedings of the IEEE/CVF Conference on Computer Vision and
  Pattern Recognition}, pages 12475--12486, 2021.

\bibitem{newell2020useful}
Alejandro Newell and Jia Deng.
\newblock How useful is self-supervised pretraining for visual tasks?
\newblock In {\em Proceedings of the IEEE/CVF Conference on Computer Vision and
  Pattern Recognition}, pages 7345--7354, 2020.

\bibitem{noroozi2016unsupervised}
Mehdi Noroozi and Paolo Favaro.
\newblock Unsupervised learning of visual representations by solving jigsaw
  puzzles.
\newblock In {\em European conference on computer vision}, pages 69--84.
  Springer, 2016.

\bibitem{noroozi2017representation}
Mehdi Noroozi, Hamed Pirsiavash, and Paolo Favaro.
\newblock Representation learning by learning to count.
\newblock In {\em Proceedings of the IEEE International Conference on Computer
  Vision}, pages 5898--5906, 2017.

\bibitem{oord2018representation}
Aaron van~den Oord, Yazhe Li, and Oriol Vinyals.
\newblock Representation learning with contrastive predictive coding.
\newblock {\em arXiv preprint arXiv:1807.03748}, 2018.

\bibitem{pan2021videomoco}
Tian Pan, Yibing Song, Tianyu Yang, Wenhao Jiang, and Wei Liu.
\newblock Videomoco: Contrastive video representation learning with temporally
  adversarial examples.
\newblock In {\em Proceedings of the IEEE/CVF Conference on Computer Vision and
  Pattern Recognition}, pages 11205--11214, 2021.

\bibitem{patrick2020multi}
Mandela Patrick, Yuki~M Asano, Polina Kuznetsova, Ruth Fong, Joao~F Henriques,
  Geoffrey Zweig, and Andrea Vedaldi.
\newblock Multi-modal self-supervision from generalized data transformations.
\newblock {\em arXiv preprint arXiv:2003.04298}, 2020.

\bibitem{qian2021spatiotemporal}
Rui Qian, Tianjian Meng, Boqing Gong, Ming-Hsuan Yang, Huisheng Wang, Serge
  Belongie, and Yin Cui.
\newblock Spatiotemporal contrastive video representation learning.
\newblock In {\em Proceedings of the IEEE/CVF Conference on Computer Vision and
  Pattern Recognition}, pages 6964--6974, 2021.

\bibitem{recasens2021broaden}
Adrià Recasens, Pauline Luc, Jean-Baptiste Alayrac, Luyu Wang, Ross Hemsley,
  Florian Strub, Corentin Tallec, Mateusz Malinowski, Viorica Patraucean,
  Florent Altché, Michal Valko, Jean-Bastien Grill, Aäron van~den Oord, and
  Andrew Zisserman.
\newblock Broaden your views for self-supervised video learning, 2021.

\bibitem{soomro2012ucf101}
Khurram Soomro, Amir~Roshan Zamir, and Mubarak Shah.
\newblock Ucf101: A dataset of 101 human actions classes from videos in the
  wild.
\newblock {\em arXiv preprint arXiv:1212.0402}, 2012.

\bibitem{srivastava2015unsupervised}
Nitish Srivastava, Elman Mansimov, and Ruslan Salakhudinov.
\newblock Unsupervised learning of video representations using lstms.
\newblock In {\em International conference on machine learning}, pages
  843--852. PMLR, 2015.

\bibitem{sun1906contrastive}
Chen Sun, Fabien Baradel, Kevin Murphy, and Cordelia Schmid.
\newblock Contrastive bidirectional transformer for temporal representation
  learning. 2019a.
\newblock {\em URL http://arxiv. org/abs}, 1906.

\bibitem{sun2017revisiting}
Chen Sun, Abhinav Shrivastava, Saurabh Singh, and Abhinav Gupta.
\newblock Revisiting unreasonable effectiveness of data in deep learning era.
\newblock In {\em Proceedings of the IEEE international conference on computer
  vision}, pages 843--852, 2017.

\bibitem{tian2020contrastive}
Yonglong Tian, Dilip Krishnan, and Phillip Isola.
\newblock Contrastive multiview coding.
\newblock In {\em Computer Vision--ECCV 2020: 16th European Conference,
  Glasgow, UK, August 23--28, 2020, Proceedings, Part XI 16}, pages 776--794.
  Springer, 2020.

\bibitem{vondrick2016anticipating}
Carl Vondrick, Hamed Pirsiavash, and Antonio Torralba.
\newblock Anticipating visual representations from unlabeled video.
\newblock In {\em Proceedings of the IEEE conference on computer vision and
  pattern recognition}, pages 98--106, 2016.

\bibitem{wang2019self}
Jiangliu Wang, Jianbo Jiao, Linchao Bao, Shengfeng He, Yunhui Liu, and Wei Liu.
\newblock Self-supervised spatio-temporal representation learning for videos by
  predicting motion and appearance statistics.
\newblock In {\em Proceedings of the IEEE/CVF Conference on Computer Vision and
  Pattern Recognition}, pages 4006--4015, 2019.

\bibitem{wang2020self}
Jiangliu Wang, Jianbo Jiao, and Yun-Hui Liu.
\newblock Self-supervised video representation learning by pace prediction.
\newblock In {\em European conference on computer vision}, pages 504--521.
  Springer, 2020.

\bibitem{wang2015unsupervised}
Xiaolong Wang and Abhinav Gupta.
\newblock Unsupervised learning of visual representations using videos.
\newblock In {\em Proceedings of the IEEE international conference on computer
  vision}, pages 2794--2802, 2015.

\bibitem{wang2019learning}
Xiaolong Wang, Allan Jabri, and Alexei~A Efros.
\newblock Learning correspondence from the cycle-consistency of time.
\newblock In {\em Proceedings of the IEEE/CVF Conference on Computer Vision and
  Pattern Recognition}, pages 2566--2576, 2019.

\bibitem{wei2018learning}
Donglai Wei, Joseph~J Lim, Andrew Zisserman, and William~T Freeman.
\newblock Learning and using the arrow of time.
\newblock In {\em Proceedings of the IEEE Conference on Computer Vision and
  Pattern Recognition}, pages 8052--8060, 2018.

\bibitem{xiao2021modist}
Fanyi Xiao, Joseph Tighe, and Davide Modolo.
\newblock Modist: Motion distillation for self-supervised video representation
  learning, 2021.

\bibitem{xiao2020should}
Tete Xiao, Xiaolong Wang, Alexei~A Efros, and Trevor Darrell.
\newblock What should not be contrastive in contrastive learning.
\newblock {\em arXiv preprint arXiv:2008.05659}, 2020.

\bibitem{xu2019self}
Dejing Xu, Jun Xiao, Zhou Zhao, Jian Shao, Di Xie, and Yueting Zhuang.
\newblock Self-supervised spatiotemporal learning via video clip order
  prediction.
\newblock In {\em Proceedings of the IEEE/CVF Conference on Computer Vision and
  Pattern Recognition}, pages 10334--10343, 2019.

\bibitem{yang2020video}
Ceyuan Yang, Yinghao Xu, Bo Dai, and Bolei Zhou.
\newblock Video representation learning with visual tempo consistency.
\newblock {\em arXiv preprint arXiv:2006.15489}, 2020.

\bibitem{zhang2016colorful}
Richard Zhang, Phillip Isola, and Alexei~A Efros.
\newblock Colorful image colorization.
\newblock In {\em European conference on computer vision}, pages 649--666.
  Springer, 2016.

\bibitem{zhu2020comprehensive}
Yi Zhu, Xinyu Li, Chunhui Liu, Mohammadreza Zolfaghari, Yuanjun Xiong, Chongruo
  Wu, Zhi Zhang, Joseph Tighe, R Manmatha, and Mu Li.
\newblock A comprehensive study of deep video action recognition.
\newblock {\em arXiv preprint arXiv:2012.06567}, 2020.

\bibitem{zhuang2019local}
Chengxu Zhuang, Alex~Lin Zhai, and Daniel Yamins.
\newblock Local aggregation for unsupervised learning of visual embeddings.
\newblock In {\em Proceedings of the IEEE/CVF International Conference on
  Computer Vision}, pages 6002--6012, 2019.

\end{thebibliography}
